\documentclass[10pt,twocolumn,letterpaper]{article}

\usepackage{wacv}              
\usepackage[accsupp]{axessibility}  

\usepackage{multirow}
\usepackage{pifont}
\newcommand{\xmark}{\ding{55}}%

\definecolor{wacvblue}{rgb}{0.21,0.49,0.74}
\usepackage[pagebackref,breaklinks,colorlinks,allcolors=wacvblue]{hyperref}

\def\wacvPaperID{1260} 
\def\confName{WACV}
\def\confYear{2026}

\title{Frequency Is What You Need: Considering Word Frequency When Text Masking Benefits Vision-Language Model Pre-training}

\author{Mingliang Liang and Martha Larson \\
Institute for Computing and Information Sciences, Radboud University, Nijmegen, Netherland \\
{\tt\small \{mliang, mlarson\}@cs.ru.nl}
}

\begin{document}
\maketitle
\begin{abstract}
Vision Language Models (VLMs) can be trained more efficiently if training sets can be reduced in size.
Recent work has shown the benefits of masking text during VLM training using a variety of strategies (truncation, random masking, block masking and syntax masking)
and has reported syntax masking as the top performer.
In this paper, we analyze the impact of different text masking strategies on the word frequency in the training data, and show that this impact is connected to model success.  
This finding motivates Contrastive Language-Image Pre-training with Word Frequency Masking (CLIPF), our proposed masking approach, which directly leverages word frequency.
Extensive experiments demonstrate the advantages of CLIPF over syntax masking and other existing approaches,
particularly when the number of input tokens decreases.
We show that not only CLIPF, but also other existing masking strategies, outperform syntax masking when enough epochs are used during training, a finding of practical importance for selecting a text masking method for VLM training. 
Our code is available online.
\end{abstract}

\section{Introduction}
\label{sec:intro}

Vision-Language Models (VLMs) learn embeddings that represent both visual and textual modalities simultaneously, and are known for the ease with which their strong performance transfers from one domain to another~\cite{DeViSE2013_Frome,mahajan2018exploring,radford2021clip,scaling2021Jia,mu2021slip,desai2021virtex,zhang2022contrastive,li2023flip}.
CLIP (Contrastive Language-Image Pre-
training)~\cite{radford2021clip,gabriel2021openclip,zhai2022lit} is a VLM that is trained on a very large number of image-text pairs collected online, where the text is considered a ``natural language'' annotation for the image.

However, pre-training CLIP models typically requires thousands of GPU-days~\cite{radford2021clip,gabriel2021openclip}.
To accelerate the pre-training process,
current work makes use of 
masking strategies for image and text tokens when pre-training CLIP models~\cite{li2023flip,yang2023aclip,li2023clipa}. 
In addition to the speed-up, masking makes it possible to use large batch sizes, which enhances the benefits of contrastive learning and has been considered as a form of regularization, reducing the risk of overfitting~\cite{li2023flip}.
For these reasons, despite the loss of some information from the original image and text, masking strategies have the potential to improve model performance.


\begin{figure}[!t]
    \centering
    \includegraphics[width=1.0\linewidth]{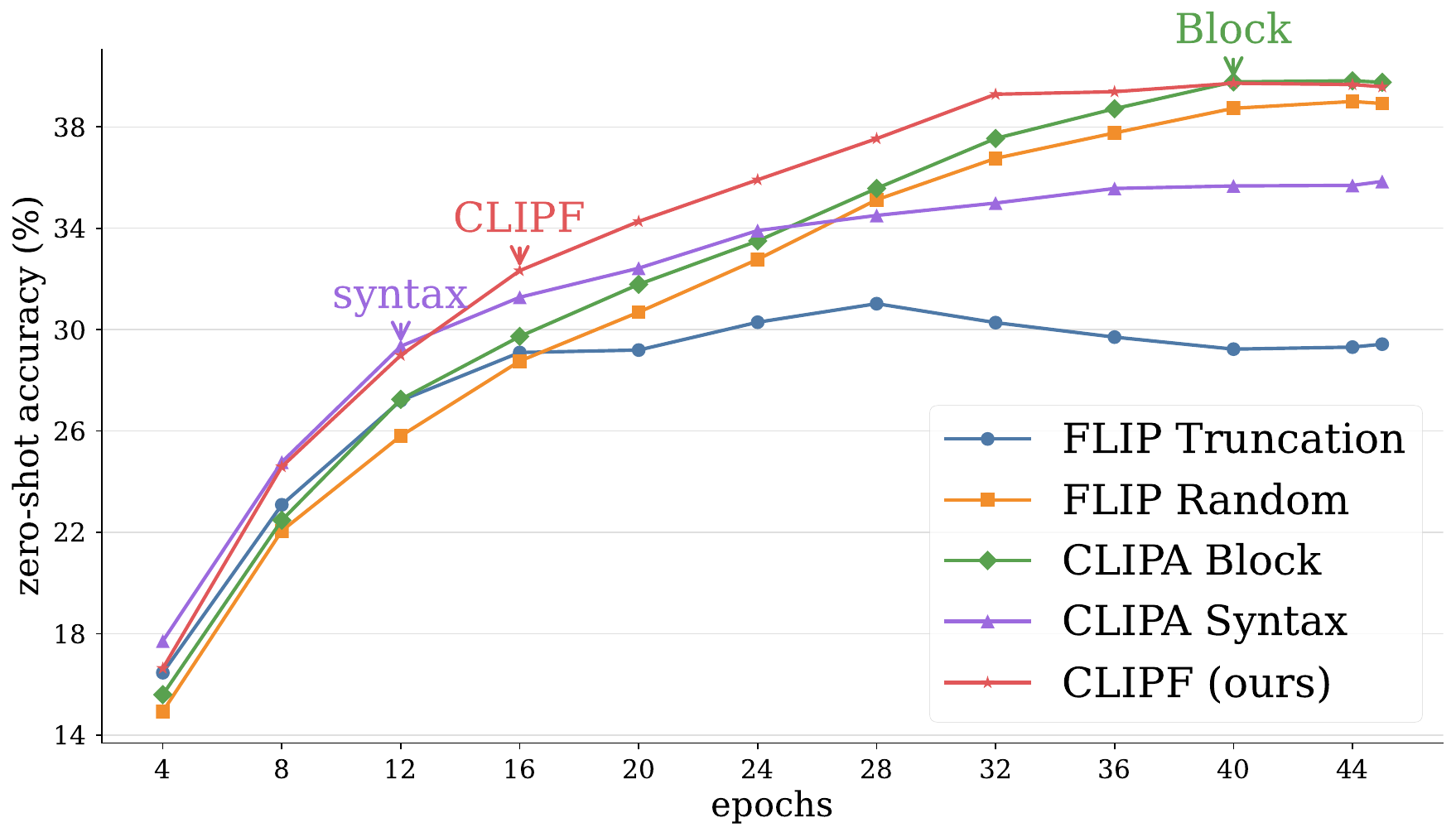}
    \caption{Zero-shot classification accuracy on ImageNet-1K using models trained on CC12M to which different text masking strategies have been applied.
    The backbone of the image encoder is ViT-B/16. The text masking ratio is 75\%.
    We also use image masking (75\%) to speed up pre-training.
    At each point, we apply an additional epoch of training on the full, unmasked data.
    }
    \label{fig:learn_curved_af}
\end{figure}

Currently, the state-of-the-art in text masking is represented by CLIPA~\cite{li2023clipa}. 
The authors compare two strategies originally introduced by FLIP~\cite{li2023flip}, truncation and random word selection, and two new approaches, random block selection and syntax-based selection, i.e., masking on the basis of part-of-speech (POS) information.
In~\cite{li2023clipa}, syntax masking is found to deliver the highest performance.

In this paper, we carry out an analysis of word frequency in the training data before and after text masking and discover a connection between the impact of a masking strategy on the training data and the success of a VLM.
As shown in \cref{fig:learn_curved_af}, our experiments reveal that syntax masking falls behind other text masking strategies (random and block) as the number of epochs increases.
Our experiments establish two important conclusions: First, syntax masking is \emph{not} the best performing text masking approach, updating the state-of-the-art finding in~\cite{li2023clipa}. 
Second, considering word frequency allows us to explain why random and block masking outperform truncation and syntax masking.

This analysis motivates Contrastive Language-Image Pre-training with Word Frequency Masking (CLIPF), a new text masking strategy that directly leverages word frequency proposed in this paper.
Because CLIPF leverages word-frequency directly, it does not need to expend compute on POS tagging, necessary for syntax masking.
~\cref{fig:learn_curved_af} shows that CLIPF outperforms syntax masking.
CLIPF is inspired by earlier training of text models, that has leveraged frequency masking~\cite{mikolov2013distributed}.
The key algorithmic novelty of CLIPF is the use of frequency-based word scores to prioritize words for masking, rather than to impose a hard decision on word removal.

An important aspect of ~\cref{fig:learn_curved_af} is the development of performance over epochs.
In contrast to the experiments in FLIP~\cite{li2023flip} and CLIPA~\cite{li2023clipa}, we continue training until additional epochs yield only minimal gains.
\cref{fig:learn_curved_af} demonstrates that our CLIPF approach is useful because it is the top performer in the range of 16-40 epochs, typically used in practice (e.g., OpenCLIP~\cite{gabriel2021openclip} trains on 30 epochs.)

In sum, our paper makes the following contributions:
\begin{itemize}
    \item We show that considering the impact of text masking on word frequency in the training data differentiates between successful and less successful text masking strategies.
    \item We introduce Contrastive Language-Image Pre-training with Word Frequency Masking (CLIPF), which leverages frequency directly during masking, and perform an in-depth quantitative comparison among existing masking strategies and CLIPF.
    \item  We point out the importance of the number of training epochs when comparing text masking strategies.
\end{itemize}
Our quantified evaluation among existing alternative text masking approaches is carried out at two scales.
A large-scale experiment on the LAION-400M~\cite{schuhmann2021laion400m} data set reproduces the CLIPA results and shows CLIPF outperforming syntax masking.
A large number of smaller scale experiments on the CC3M~\cite{sharma2018cc3m} and CC12M~\cite{changpinyo2021cc12m} data sets, which are widely used for VLM research~\cite{gabriel2021openclip,li2022DeCLIP,mu2021slip,yang2023aclip,li2022blip}, demonstrate that consideration of frequency is important in two ways.
First, the frequency-based approach CLIPF emerges an excellent choice for text masking, especially as the number of input tokens decreases.
Second, the text masking approach should be selected specifically for the application scenario (i.e., data set, input size, number of training epochs).
However, in general masking strategies that disproportionately impact word frequency (truncation and syntax) should be avoided and strategies that protect relative word frequencies (random, block and CLIPF) should be preferred.
Our code is available online.\footnote{https://github.com/MingliangLiang3/CLIPF}



\section{Related work}
\label{sec:related work}

\subsection{Vision-Language Models} 
Vision-language models (VLMs) learn image embeddings from large-scale natural language supervision,
which is important for making models transferable.
CLIP~\cite{radford2021clip} is the leading example of a VLM widely-used in practice, and other examples are ALIGN, SigLIP, and FLIP~\cite{radford2021clip,scaling2021Jia,zhai2023siglip,li2023flip}, 
CLIP utilizes an image encoder and a text encoder to generate the representation of the image and text, and learn the joint representation by contrastive learning~\cite{radford2021clip,oord2018InfoNCE}.
Because CLIP does not release its pre-training dataset, OpenCLIP pre-trained models on several large-scale public datasets~\cite{changpinyo2021cc12m,thomee2016yfcc100m,schuhmann2021laion400m,schuhmann2022laionb} to reproduce CLIP.


Although CLIP and OpenCLIP are pre-trained on very large-scale datasets, recent works like DeCLIP~\cite{li2022DeCLIP} and SLIP~\cite{mu2021slip} demonstrate that Vision-language Models (VLMs) can be trained efficiently with smaller datasets, enabling researchers with limited computational resources to contribute. 
Moreover, masking image patches or reducing image resolution during VLM pre-training also substantially 
speed up pre-training
while maintaining or enhancing performance, as shown by techniques like FLIP, RECLIP, A-CLIP and CLIPA~\cite{li2023flip,li2023reclip,yang2023aclip,li2023clipa}.

\subsection{Text masking for VLM training} 


Masking text presents another viable option to speed up and enhance the pre-training of VLMs, and is particularly effective alongside strategies that employ a high ratio of image patch masking~\cite{li2023flip,li2023clipa}. 
FLIP further explored text masking techniques, namely, random and truncation masking (referred to as ``prioritized''), to enhance efficiency~\cite{li2023flip}.
While random text masking involves masking text arbitrarily, truncation masking in FLIP strategically eliminates padding tokens and retains the words that preceded them.
The study of text masking for VLMs got off to a slow start, 
since~\cite{li2023flip} found that text masking promoted efficiency, but deteriorated performance.

Subsequently, CLIPA~\cite{li2023clipa} introduced block masking and syntax masking.
As already mentioned in~\cref{sec:intro}, 
CLIPA~\cite{li2023clipa} shows that syntax masking achieves the best performance when applying text masking.
CLIPA syntax masking prioritizes the retention of nouns, followed by adjectives and then other words.
In~\cite{li2023clipa}, syntax masking outperforms the truncation, random, and block approaches.
However,~\cite{li2023clipa} experiments only with a fixed number of epochs, and does not test performance as the number of training epochs increases. 
An important contrast with our work, as previously mentioned, is our attention to the number of training epochs. 
Further, CLIPF, as we will see, counters the danger of giving overly high priority to nouns, leading to overfitting, and also avoids overly suppressing words of the POS category ``Other''. 

As mentioned in the introduction, CLIPF is inspired by earlier training of text models~\cite{mikolov2013distributed}, which selected words by frequency.
An initial attempt was made to transfer this idea to CLIP training in a short workshop paper~\cite{liang2023swclip} published at the same time as CLIPA.
The algorithmic novelty of our CLIPF strategy compared to the frequency masking in ~\cite{mikolov2013distributed} and~\cite{liang2023swclip} is the use of the word masking probability score for prioritization, based on the insight that the hard threshold must be dropped.
In short, in order to exploit frequency information effectively, each training text needs to fill as many available input tokens as possible, as is done in CLIPF.

\section{Considering Word Frequency For Masking}
\label{sec:freq}
In this section, we introduce existing text masking strategies and carry out an analysis on their impact on word frequency in the training data, which reveals why random and block masking outperform syntax and truncation masking.

\subsection{Existing Text Masking Strategies}
\label{ref:clipameth}

Here, we define text masking and provide descriptions of the existing text masking strategies introduced by FLIP~\cite{li2023flip} and CLIPA~\cite{li2023clipa}.

\textbf{Text Masking:} Consider a text $T_i$ consisting of $n$ words $\{w_1, w_2, w_3, \ldots, w_n\}$.
Text masking involves selecting a subset of these words to create a new version of the text, denoted as
\begin{equation}
    T'=f(T\{w_1, w_2, w_3, \ldots, w_m\}), 
\end{equation}
\label{equ:text}
In the context of VLMs, the result of text masking is a string of tokens:
\begin{equation}
    T'=f(T\{t_1, t_2, t_3, \ldots, t_k\}), 
\end{equation}
\label{equ:token}
where $k$ is the length of the input and is determined by the masking ratio.
The text is then tokenized, and $k$ tokens are chosen for use in training. 


\textbf{Truncation Masking:} In truncation masking, the first $k$ tokens in text $T$ are used as the input.


\textbf{Random Masking:} In  random text masking, the function $f$ operates to randomly select tokens from the text $T$.
Each token $t_i$ has an equal probability of being masked, independently of its position or frequency in the text.

\textbf{Block Masking:}
Block masking is similar to truncation masking but starts at a random token index $j$ within the text $T$. 
If $T$ contains more tokens than there are input tokens, we choose a random starting index $j$ that is less than the number of the text tokens minus the number of the input tokens $k$.

\textbf{Syntax Masking:} To implement syntax masking~\cite{li2023clipa}, each word $w_i$ in the text $T$ is first tagged using a natural language processing tool like NLTK~\cite{bird2009nltk} to identify its part of speech (POS). 
The words are then sorted according to their syntax role. 
Priority is given to nouns, followed by adjectives, verbs and then words belong to other POS categories.
The first $k$ tokens are used as the input.

\begin{figure}[!t]
    \centering
    \includegraphics[width=1.0\linewidth]{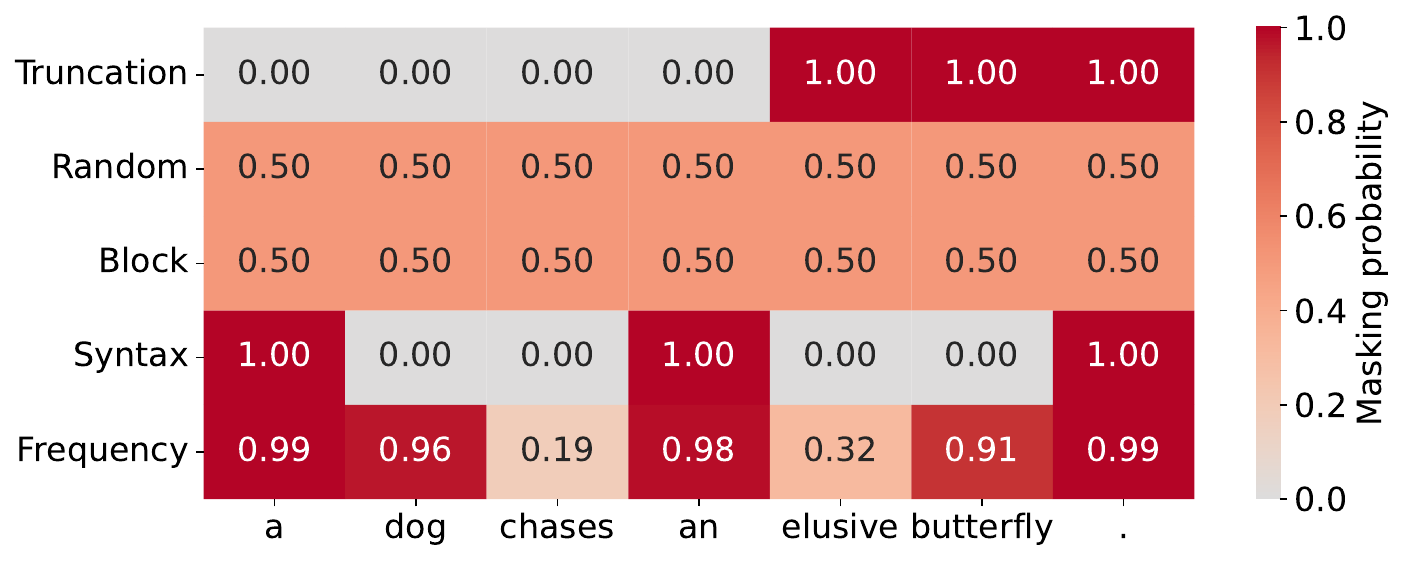}
    \caption{Comparison of word masking probabilities for various methods.
    In this example, we keep four words for truncation and syntax.
    Numbers indicate the probability of a word being masked.
    \textbf{Truncation} keeps the first 4 words; 
    \textbf{Random} and \textbf{Block} mask each word with a 50\% probability;
    \textbf{Syntax} prioritizes retaining nouns, followed by adjectives, then others;
    \textbf{Frequency} masks words based on the frequency of the words.
    }
    \label{fig:text_masking_case}
\end{figure}

An example that illustrates the way that the different forms of masking impact a text in the training set is provided in~\cref{fig:text_masking_case}.

\subsection{Impact of masking on word frequency}

\begin{figure}[!t]
    \centering
    \includegraphics[width=1.0\linewidth]{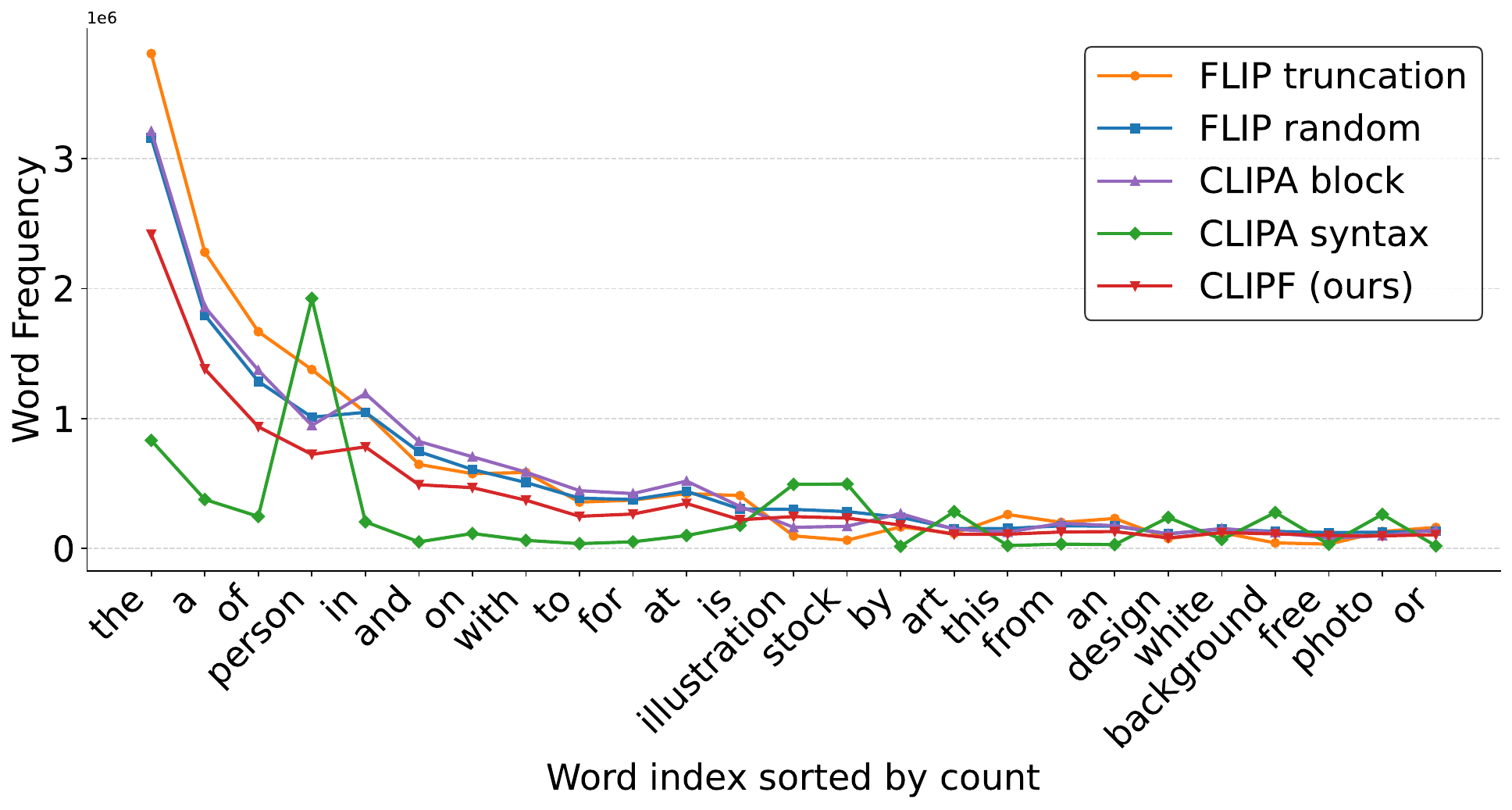}
    \caption{The figure illustrates the distribution of top-25 words in the text before and after applying various text masking strategies. We set the text length after text masking to 6.
    The x-axis represents the word index, which is sorted by counts of the original data, and the y-axis shows the word frequency.
    The dataset used is CC12M and the value of $t$ of~\cref {equ:sub} is set to $10^{-6}$.
    We remove special characters from the vocabulary. 
    The figure with more words is provided in the supplementary material.}
    \label{fig:word_frequency_distribution}
\end{figure}

Existing literature reports the impact of text masking in terms of the percentage of words or tokens retained in the training data. 
Seen from the word-count perspective, the impact is the same: for any text masking approach, the proportion of words removed controls the computational savings.
However, to determine the reasons why some text masking approaches are better than others, it is necessary to look beyond overall word count.

In this section, we carry out an analysis that goes beyond word count and considers the impact of text masking on the frequency of words occurring in the training set.
We plot word-frequency curves, used in linguistics, for the four existing approaches to text masking and display the head of the curve (the first 25 most frequent words) in~\cref{fig:word_frequency_distribution}. 
The curves generally follow the long-tail distribution expected of natural language and predicted by Zipf's Law~\cite{kingsley1932zipf}.
Further, linguistic theory leads us to expect that the top-frequency words are function words (e.g., articles and prepositions), which is indeed what we observe here.
However, looking at individual words in~\cref{fig:word_frequency_distribution}, we see aberrations.
Two observations are important: First, for truncation masking the four most frequent words are markedly more frequent than with the other forms of masking.
These are function words (``the'', ``a'', and ``of'') and one noun (``person''), which reflects the fact that sentences tend to start with these words.
Second, for syntax masking certain nouns form peaks, meaning that they are more frequently retained with syntax masking than with other forms of masking.


Given the design of syntax masking, it is unsurprising that the words that are disproportionately retained belong to the POS category Noun. 
We dive deeper into the impact of existing masking approaches by examining the word distribution over POS categories. 
~\cref{tab:syntax_percentage} illustrates statistics for four major part-of-speech classes, Nouns, Adjectives, Verbs and Other (predominantly function words).
\begin{table}[!t]
    \centering
    \resizebox{\linewidth}{!}{%
    \begin{tabular}{lrrrrrr}
    \toprule
    \textbf{Masking}  & \textbf{NN (\%)} & \textbf{JJ (\%)} & \textbf{VB (\%)} & \textbf{OTHER (\%)} \\
    \midrule
    Before masking    & 50.30\% & 4.98\% & 5.18\% & 39.55\% \\ \hline
    Truncation        & 51.72\% & 5.17\% & 5.91\% & 37.20\% \\
    Random            & 51.21\% & 4.82\% & 5.04\% & 38.93\% \\
    Block             & 49.37\% & 4.74\% & 5.24\% & 40.66\% \\
    Syntax            & 88.53\% & 2.69\% & 2.92\% & 5.87\%  \\ \hline
    CLIPF             & 60.43\% & 5.06\% & 6.20\% & 28.24\% \\ 
    \bottomrule
    \end{tabular}
    }
    \caption{Part-of-speech (POS) distribution: The percentage of nouns (NN), adjectives (JJ), verbs (VB) and Other POS categories before and after text masking, calculated on CC12M, retaining 6 words per text.}
    \label{tab:syntax_percentage}
\end{table}
We observe that the distribution for truncation, random, and block masking is comparable to the distribution before masking. 
However, for syntax masking, the number of words in the Noun category is extremely high and the number of words in the Other category is extremely low. 

The prioritization of the POS category Noun and the deprioritization of Other has intuitive appeal.
Although not explicitly mentioned, we assume that this appeal motivated CLIPA~\cite{li2023clipa} in their design of syntax masking.
Nouns are likely to be the name of an object directly depicted in an image, and can form a better bridge between image and text than words in the categories Adjectives or Verb.
Further, words in the category of Other, being function words, are often classical stop words that are removed in information retrieval applications.
Our results (\cref{fig:learn_curved_af}) show that this intuition does not directly lead to good performance and must be revisited.

In short, our analysis reveals the importance of considering word frequency when text masking. We have seen that text masking approaches that do not disproportionately impact the frequency of certain words (i.e., random and block) outperform text masking strategies that do (i.e., truncation and syntax). 
The word frequency changes caused by using sentence-initial words (truncation) or emphasizing/de-emphasizing certain POS categories (syntax) hold back performance as the number of epochs increases.
This observation motivates the importance of a text masking approach that is directly maintains relative word frequency, which we introduce in the next section.


\section{Word Frequency Masking (CLIPF)}
\label{ref:clipfmeth}
Contrastive Language-Image Pre-training with Word Frequency Masking (CLIPF) removes a word $w_i$ from the training data according to its masking probability score, $P(w_i)$, which is calculated on the basis of the word's frequency as: 

\begin{equation}
    P(w_i) = 1 - \sqrt{\frac{t}{f(w_i)}}
    \label{equ:sub}
\end{equation}

Word frequency, $f(w_i)$, is the number of times the word appears in the training data divided by the total number of words in the training data.
Eq.~\ref{equ:sub} scales $f(w_i)$, resulting in a much higher masking probability for frequent words than infrequent ones. 
It also preserves the overall ranking of word frequencies.

The aggressiveness with which high frequency words are masked is controlled in Eq.~\ref{equ:sub} by the threshold $t$.
CLIPF enjoys the benefit of being relatively insensitive to the parameter $t$, meaning that the parameter does not have to be precisely tuned.
The influence of $P(w_i)$ on $t$ is illustrated by a plot in the supplemental material, which also includes an experimental demonstration that the CLIPF is relatively insensitive to the choice of $t$. 

The basic form of Eq.~\ref{equ:sub} has been validated previously in the literature, having been introduced by~\cite{mikolov2013distributed} for the purpose of text model training, which inspired CLIPF.
Specifically,~\cite{mikolov2013distributed} and SW-CLIP~\cite{liang2023swclip} (which attempted to transfer~\cite{mikolov2013distributed} to CLIP training) discard words whose $P(w_i)$ exceeds a cutoff. 
If there are not enough low $P(w_i)$ words in a training text to fill in the input, these approaches use padding.
CLIPF does away with this cutoff, and instead fills the input with words in order of $P(w_i)$.
Because CLIPF uses higher-$P(w_i)$ words rather than padding it can outperform SW-CLIP, as shown in the supplementary material.






\section{Performance of CLIPF}
In this section, we discuss the performance of CLIPF on the basis of a reproduction of the CLIPA paper (~\cref{LAION-400M}) with a model trained on LAION-400M~\cite{schuhmann2021laion400m} as well as the initial experiment in \cref{fig:learn_curved_af} with models trained on CC3M~\cite{sharma2018cc3m} and  CC12M~\cite{changpinyo2021cc12m}.


\subsection{Revisiting CLIPA}
\label{LAION-400M}
We carry out a reproduction of the original CLIPA experiment on LAION-400M~\cite{schuhmann2021laion400m}.
We remain as faithful as possible to the original setup, with two exceptions.
First, not all 400M URLs from LAION-400M are still available online, and we train on the 298M pairs that are available. 
Second, we reduce the computational expense of the experiment by using a smaller image size (112x112) and 50\% image masking.
We employ a ViT-B/16 image encoder with global average pooling and learnable positional embeddings.
As with CLIPA, we fine-tune for 0.4 epochs on the full data after training on the masked data.
More details about the experimental setup are available in the supplementary material.

\begin{figure}
    \centering
    \includegraphics[width=0.8\linewidth]{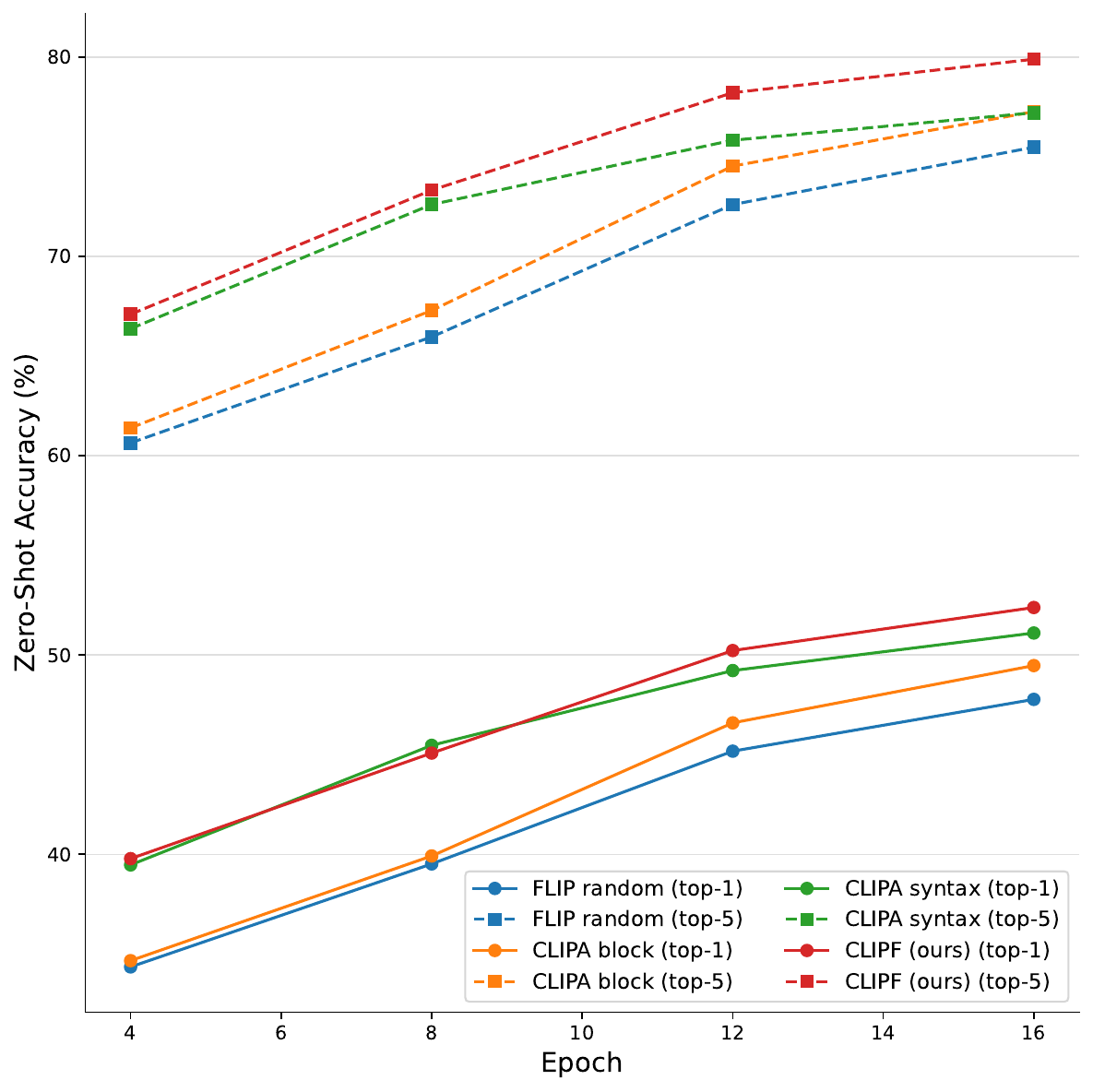}
    \caption{Zero-shot classification accuracy on ImageNet-1K using models trained on LAION-400M to which different text masking strategies have been applied.
    The backbone of the image encoder is ViT-B/16. 
    We pre-trained the model using 25 image tokens and 4 text tokens for 16 epochs.
    At each point, we apply an additional 0.4 epoch of training on the full, unmasked data.
    }
    \label{fig:laion400m}
\end{figure}

Results for zero-shot classification on ImageNet-1K are shown in \cref{fig:laion400m} for both top-1 accuracy (bottom) and top-5 accuracy (top) for 4-16 epochs of training on the masked data.
We observe that CLIPF outperforms syntax masking. 

The original CLIPA paper established that truncation masking was not a good performer, and we do not test it here.
However, we find that other aspects of the conclusions of CLIPA concerning text masking are limited.
Specifically, in order to observe this difference it is necessary to continue training until the increase in performance starts to level off.
The original CLIPA paper was trained with only 6.4 epochs, and here we train to 16.
The additional epochs allow us to observe block masking and random masking moving towards syntax masking as the number of epochs increases, with block masking on par with syntax masking with respect to top-5 accuracy at 16 epochs.

Comparing \cref{fig:laion400m} with the first part of \cref{fig:learn_curved_af}, we see that the relative levels of performance of syntax, block, random and CLIPF are the same. 
This comparability gives us confidence that we can study text masking with smaller data sets, which allows us to train for a large number of epochs using different comparative conditions, and we carry out the remaining experiments on data sets that require less computational expenditure.

\subsection{Advantages of CLIPF}
\label{sec:advantages}
We report on the performance of CLIPF vs. the original CLIP with an experiment carried out on CC3M~\cite{sharma2018cc3m} and  CC12M~\cite{changpinyo2021cc12m}.
Not all text-image pairs were available online and our final data sets consist of 2.7M pairs from CC3M and 9.3M pairs from CC12M.
For CLIPF, we fine-tune for one epoch on the full data after training on the masked data.
We employ the same ViT-B/16 image encoder, but we use a 224×224 image input size.
Further details about the experimental setup are available in the supplementary material.
The results in \cref{tab:text_masking_more_image_token} confirm that CLIPF is able to outperform CLIP.
``Image tokens'' and ``text tokens'' reflect the amount of image and text processing during pre-training and this performance gain comes with the computational advantage of using half the number of image tokens and one quarter of the number of text tokens. 
\begin{table}[!t]
\centering
\resizebox{\linewidth}{!}{%
    \begin{tabular}{l|l|c|c|cc|cc}
    \toprule
    \multirow{2}{*}{\textbf{Models}} & \multirow{2}{*}{Masking} & \multirow{2}{*}{\begin{tabular}[c]{@{}c@{}}Image \\ Tokens\end{tabular}} & \multirow{2}{*}{\begin{tabular}[c]{@{}c@{}}Text \\ Tokens\end{tabular}} &  \multicolumn{2}{c|}{CC3M} & \multicolumn{2}{c}{CC12M} \\ 
                                    & &  & & pre-train & fine-tune & pre-train & fine-tune \\ 
    \midrule
    \textbf{CLIP} & \multirow{1}{*}{\textbf\xmark} & 197 & 32 & 18.6 & \xmark & 36.6 & \xmark \\ \hline
    \textbf{CLIPF}
                & frequency   & 98   & 8   & \textbf{19.6} & \textbf{18.7} & \textbf{39.8}  & \textbf{41.0}   \\ \bottomrule
    \end{tabular}
}
\caption{\textbf{Comparison of  zero-shot classification on ImageNet-1K between CLIP and CLIPF using 98 image tokens and 8 text tokens. } Reported as top-1 accuracy.
All models use a ViT-B/16 image encoder backbone.
They are pre-trained on \textbf{CC3M} and \textbf{CC12M} for 30 epochs (pre-train). CLIPF is  fine-tuned for an additional epoch on all (no masking) data (fine-tune).
}
\label{tab:text_masking_more_image_token}
\end{table}

When text masking in an application scenario, choices need to be made, as already mentioned in~\cref{sec:intro}.
As seen in~\cref{fig:learn_curved_af}, the number of epochs matters. 
Here, we choose to train with 30 epochs, following~\cite{gabriel2021openclip}.
The aggressiveness of the masking used will be determined by the computational resources available in the application scenario.
Here, we used 50\% image masking and in our further experiments, we adopt 25\% image masking for greater speed. 
In the next section, we carry out a quantified evaluation of all existing text masking approaches, comparing them to CLIPF.
This evaluation provides the information necessary to support the choices of text masking approaches, including the number of text input tokens, for use in application scenarios.

\section{Comparison of Text Masking Strategies}
In this section, we carry out a systematic evaluation of the existing text masking approaches from FLIP~\cite{li2023flip} (random and truncation), CLIPA~\cite{li2023clipa} (block and syntax) and our CLIPF approach (frequency-based).
In a nutshell, the results support our observation that syntax and truncation masking are not productive approaches to masking and highlight the benefits of masking that directly leverages frequency information, i.e., our CLIPF. 
These experiments are carried out on CC3M~\cite{sharma2018cc3m} and  CC12M~\cite{changpinyo2021cc12m} with the same experimental setup as in \cref{sec:advantages}, using 30 epochs, as previously mentioned, following OpenCLIP~\cite{gabriel2021openclip}. 
Following prior evaluations of CLIP models~\cite{radford2021clip,li2023flip}, we evaluate the models' zero-shot performance on ImageNet-1k (\cref{sec:0shotImageNet}), zero-shot robustness (\cref{sec:0robust}), zero-shot image-text retrieval (\cref{sec:0retrieval}) and zero-shot classification on 25 additional data sets (\cref{sec:add_data}).
Further details of the experimental setup are in the supplementary material.


\subsection{Zero-shot Performance on ImageNet-1k} 
\label{sec:0shotImageNet}
\cref{tab:text_masking} presents results for zero-shot classification on ImageNet-1K, which complement the ImageNet-1K results already presented in \cref{fig:learn_curved_af}.
The experiments compare different masking approaches and different masking ratios, indicated in the table with the number of input text tokens (Text Tokens column). 
Models are pre-trained for 30 epochs (pre-train column) and then fine-tuned for one epoch (fine-tune column) with the full, unmasked data.
Note that since the average text length in CC3M is less than 16 tokens, the benefits of fine-tuning a model pre-trained with 16 text tokens are limited.

From \cref{tab:text_masking}, we observe the ability of fine-tuned text-masking strategies to outperform FLIP~\cite{li2023flip} in many cases (image masking alone), while using fewer text input tokens.
This result is important since~\cite{li2023flip} states that text masking does not contribute to improvement in model performance. 
We attribute the fact that we find a gain to the limited number of epochs (6.4) with which~\cite{li2023flip} tested text masking.

From \cref{tab:text_masking}, we also observe that CLIPF outperforms the other masking matching approaches in nearly all cases across text masking ratios (input sizes), as reflected by the bolded accuracy scores.
Fine-tuned CLIPF delivers good performance both on CC3M and also the larger CC12M.
On CC12M, fine-tuned CLIPF even outperforms the original CLIP (no masking) with a 75\% masking ratio (8 input tokens).
Note that at a text masking ratio of 75\% (corresponding to 8 tokens), the masking approaches are 1.3 $\times$ faster than FLIP and 4.0 $\times$ faster than CLIP.

Truncation performs poorly, consistent with the observations of CLIPA~\cite{li2023clipa}.
However, in contrast to~\cite{li2023clipa}, we observe that syntax masking is a poor performer. 
In contrast, approaches that do not disproportionately impact the frequencies of words (random and block) perform more strongly.





Finally, we consider the pattern of the text masking approaches as the masking ratio increases, i.e., as we decrease the input size from 16 to 4 tokens.
Here, we focus on the results of the fine-tuned model.
Moving from 16 to 8 input tokens benefits most approaches.
The benefit derives from the fact that the reduction of the number of input tokens allows for an increase in the batch size.
As we reduce the number of tokens from 8 to 6 to 4, the gap between fine-tuned CLIPF and the next best masking strategy widens.
We see that compared to existing text matching approaches, CLIPF is better positioned to deliver benefits as the number of input tokens decreases.

\begin{table}[!t]
\centering
\resizebox{\linewidth}{!}{%
    \begin{tabular}{l|l|c|c|cc|cc}
    \toprule
    \multirow{2}{*}{\textbf{Models}} & \multirow{2}{*}{Masking} & \multirow{2}{*}{\begin{tabular}[c]{@{}c@{}}Image \\ Tokens\end{tabular}} & \multirow{2}{*}{\begin{tabular}[c]{@{}c@{}}Text \\ Tokens\end{tabular}} &  \multicolumn{2}{c|}{CC3M} & \multicolumn{2}{c}{CC12M} \\ 
                                    & &  & & pre-train & fine-tune & pre-train & fine-tune \\ 
    \midrule
    \textbf{CLIP} & \multirow{1}{*}{\textbf\xmark} & 197 & 32 & 18.6 & \xmark & 36.6 & \xmark \\ \hline
    \textbf{FLIP} & \xmark & 49 & 32 & 14.1 & 14.2 & 32.0 & 33.7 \\ \hline
    
    \multirow{2}{*}{\textbf{FLIP}}    
                & truncation  & \multirow{5}{*}{49} & \multirow{5}{*}{16} & 13.8 & 13.8 & 32.8 & 32.8 \\ 
                & random      &  &  & 13.9 & 13.9 & 33.7 & 34.3 \\ \cline{1-2}
    \multirow{2}{*}{\textbf{CLIPA}} 
                & block       &  &  & 13.9 & 13.9 & 34.3 & 34.8 \\ 
                & syntax      &  &  & 13.3 & 12.8 & 32.2 & 34.4 \\ \cline{1-2}\cline{5-8}
    \textbf{CLIPF}
                & frequency   &  &  & \textbf{14.0} & \textbf{14.0} & \textbf{35.5} & \textbf{36.0} \\ \hline
    
    \multirow{2}{*}{\textbf{FLIP}}    
                & truncation  & \multirow{5}{*}{49} & \multirow{5}{*}{8} & 10.8 & 12.0 & 25.4 & 28.4 \\ 
                & random      &  &  & \textbf{17.6} & 17.4 & 34.5 & 36.9 \\ \cline{1-2}
     \multirow{2}{*}{\textbf{CLIPA}}   
                & block       &  &  & 16.2 & 16.6 & 35.5 & 37.9 \\ 
                & syntax      &  &  & 17.2 & \textbf{17.5} & 28.5 & 35.0 \\ \cline{1-2}\cline{5-8}
    \textbf{CLIPF}
                & frequency   &  &  & 16.8 & 17.0 & \textbf{36.6} & \textbf{39.3} \\ \hline
    
    \multirow{2}{*}{\textbf{FLIP}}    
                & truncation  & \multirow{5}{*}{49} & \multirow{5}{*}{6} & 8.4  & 9.4 & 15.3 & 23.2 \\ 
                & random      &  &  & 12.8 & 17.9 & 26.9 & 34.6 \\ \cline{1-2}
    \multirow{2}{*}{\textbf{CLIPA}}   
                & block       &  &  & 12.9 & 17.0 & 28.6 & 35.9 \\ 
                & syntax      &  &  & 12.2 & 15.7 & 25.2 & 32.6 \\ \cline{1-2}\cline{5-8}
    \textbf{CLIPF}
                & frequency   &  &  & \textbf{14.4} & \textbf{18.2} & \textbf{30.3} & \textbf{37.8} \\ \hline
    
    \multirow{2}{*}{\textbf{FLIP}}    
                & truncation  & \multirow{5}{*}{49} & \multirow{5}{*}{4} & 3.8  & 8.2 & 5.3  & 19.8 \\ 
                & random      &  &  & 5.4  & 14.6 & 14.0 & 27.1 \\ \cline{1-2}
    \multirow{2}{*}{\textbf{CLIPA}}   
                & block       &  &  & 7.5  & 14.5 & \textbf{18.7} & 26.6 \\ 
                & syntax      &  &  & 8.9  & 13.0 & 14.2 & 24.6 \\ \cline{1-2}\cline{5-8}
    \textbf{CLIPF}
                & frequency   &  &  & \textbf{10.9} & \textbf{16.0} & 17.0 & \textbf{30.9} \\ \hline
    \end{tabular}
}
\caption{\textbf{Comparison of CLIP, FLIP, CLIPA and CLIPF for zero-shot classification on ImageNet-1K. } Reported as top-1 accuracy.
All models use a ViT-B/16 image encoder backbone pre-trained on \textbf{CC3M} and \textbf{CC12M} for 30 epochs with 75\% image masking for speedup (pre-train) and fine-tuned for an additional epoch on all data (fine-tune).
``Image tokens'' and ``text tokens'' reflect the amount of image and text processing during pre-training.
}
\label{tab:text_masking}
\end{table}

\subsection{Zero-shot Robustness Evaluation}
\label{sec:0robust}

\begin{table}[!t]
    \centering
    \resizebox{1.0\linewidth}{!}{
    \begin{tabular}{l|l|c|c|cccccc}
    \toprule
    \multirow{2}{*}{Models} & \multirow{2}{*}{Masking} & \multirow{2}{*}{\begin{tabular}[c]{@{}c@{}}Image \\ Tokens\end{tabular}} & \multirow{2}{*}{\begin{tabular}[c]{@{}c@{}}Text \\ Tokens\end{tabular}}
     & \multicolumn{6}{c}{ViT-B/16} \\ 
    & & & & IN-A & IN-O & IN-R & IN-S & IN-V2 & ON \\ 
    \midrule
    CLIP & \xmark & 197 & 32 & 8.97 & 37.85 & 49.11 & 25.70 & 31.48 & 24.20 \\ \hline
    FLIP & \xmark & 49 & 32 & 6.90 & 40.30 & 36.90 & 17.60 & 28.70 & 18.10 \\ \hline

    \multirow{2}{*}{FLIP} 
        & truncation & \multirow{5}{*}{49} & \multirow{5}{*}{16} & 7.10 & 37.70 & 36.40 & 17.50 & 27.40 & 17.20 \\
        & random & & & 8.05 & 38.60 & 40.00 & 20.34 & 29.61 & 20.84 \\ \cline{1-2}
    \multirow{2}{*}{CLIPA}
        & block & & & 8.05 & 39.60 & 39.17 & 19.31 & 29.70 & 19.72 \\
        & syntax & & & 8.00 & \textbf{40.70} & 39.41 & 18.61 & 29.17 & 19.96 \\ \cline{1-2}\cline{5-10}
    CLIPF & frequency & & & \textbf{8.28} & 39.10 & \textbf{40.86} & \textbf{20.48} & \textbf{30.78} & \textbf{21.15} \\ \hline

    \multirow{2}{*}{FLIP}  & truncation & \multirow{5}{*}{49} & \multirow{5}{*}{8} 
        & 6.71 & 34.05 & 37.18 & 15.99 & 24.49 & 15.98 \\
        & random & & & 9.16 & 38.15 & 43.35 & 22.53 & 31.24 & \textbf{22.17} \\  \cline{1-2}  
    \multirow{2}{*}{CLIPA}
        & block & & & 9.05 & \textbf{39.95} & 43.32 & 22.58 & 32.42 & 21.58 \\
        & syntax & & & 8.50 & 38.40 & 39.90 & 19.50 & 29.80 & 19.50 \\ \cline{1-2}\cline{5-10}
    CLIPF & frequency & & & \textbf{9.56} & 39.90 & \textbf{45.76} & \textbf{23.62} & \textbf{34.16} & 21.59 \\ \hline

    \multirow{2}{*}{FLIP}  
        & truncation & \multirow{5}{*}{49} & \multirow{5}{*}{6} & 6.70 & 26.60 & 33.20 & 13.30 & 20.20 & 14.80 \\
        & random & & & 8.10 & 36.10 & 41.80 & 21.10 & 30.20 & 19.80 \\ \cline{1-2}
    \multirow{2}{*}{CLIPA}
        & block & & & 8.32 & 38.20 & 42.99 & 21.73 & 30.82 & 19.76 \\
        & syntax & & & 7.40 & 34.10 & 39.60 & 18.50 & 28.60 & 19.80 \\ \cline{1-2}\cline{5-10}
    CLIPF & frequency & & & \textbf{8.69} & \textbf{38.55} & \textbf{45.09} & \textbf{23.45} & \textbf{32.59} & \textbf{20.04} \\ \hline

    \multirow{2}{*}{FLIP}  
        & truncation & \multirow{5}{*}{49} & \multirow{5}{*}{4} & 5.30 & 21.70 & 25.40 & 9.00 & 17.10 & 11.20 \\
        & random & & & 6.10 & 30.30 & 33.80 & 14.90 & 23.50 & 15.30 \\ \cline{1-2}
    \multirow{2}{*}{CLIPA}
        & block & & & 6.08 & 30.55 & 34.69 & 15.54 & 22.99 & 15.10 \\
        & syntax & & & 6.50 & 29.10 & 32.40 & 13.30 & 22.00 & 16.00 \\ \cline{1-2}\cline{5-10}
    CLIPF & frequency & & & \textbf{7.05} & \textbf{32.40} & \textbf{39.90} & \textbf{18.94} & \textbf{26.72} & \textbf{17.14} \\ 
    \bottomrule
    \end{tabular}}
    \caption{Zero-shot robustness evaluation. Comparison of the zero-shot accuracy performance of CLIP, FLIP, CLIPA, and CLIPF on various datasets.
    The models are pre-trained on \textbf{CC12M}~\cite{changpinyo2021cc12m} for 30 epochs with image masking (75\%) to speed up training and fine-tune the model an additional epoch without image and text masking.
    }
    \vspace{-0.3cm}
    \label{tab:imagenet}
\end{table}
    
\cref{tab:imagenet} presents results for zero-shot classification for six data sets commonly used for robustness evaluation (e.g., by~\cite{radford2021clip,li2023flip}):
ImageNet-A (IN-A)~\cite{hendrycks2021imageneta}, ImageNet-O (IN-O)~\cite{hendrycks2021imageneto}, ImageNet-R (IN-R)~\cite{hendrycks2021imagenetr}, ImageNet-Sketch (IN-S)~\cite{wang2019imagesketch}, ImageNet-V2 (IN-Vs)~\cite{recht2019imagenetv2}, and ObjectNet (ON)~\cite{barbu2019objectnet}.
The results lead to the same observations as with the zero-shot performance on ImageNet-1k in \cref{sec:0shotImageNet}.
Across different text masking ratios and across different data sets, CLIPF outperforms the other masking strategies in nearly all cases.
Truncation and syntax are generally not good performers.
Reducing the text tokens in CLIPF from 16 to 8 results in substantial performance improvements compared to the other text masking approaches across almost all datasets.
Again, we see that the benefits of CLIPF are particularly evident as the masking number of input tokens decreases. 

We also observe that CLIPF is able to improve over the FLIP baseline, without text masking in a majority of cases.
However, despite being the strongest text-masking strategy, the results reveal that it is challenging for CLIPF to improve over the original CLIP.
In only a handful of cases does it outperform the original CLIP, but might do so with more image tokens as we saw in~\cref{sec:advantages}.

\begin{table}[!t]
    \centering
    \resizebox{\linewidth}{!}{%
    \begin{tabular}{l|l|c|c|c|c|c|c}
    \toprule
       \multirow{3}{*}{Models} & \multirow{3}{*}{Masking} & \multirow{3}{*}{\begin{tabular}[c]{@{}c@{}}Image \\ Tokens\end{tabular}} & \multirow{3}{*}{\begin{tabular}[c]{@{}c@{}}Text \\ Tokens\end{tabular}} & \multicolumn{2}{c|}{Text Retrieval}    & \multicolumn{2}{c}{Image Retrieval}    \\
                            & & & & \multicolumn{1}{c|}{Flickr30k} & \multicolumn{1}{c|}{COCO} & \multicolumn{1}{c|}{Flickr30k} & \multicolumn{1}{c}{COCO} \\
                            & & & & R@1 & R@1 & R@1 & R@1 \\
                        \midrule
        CLIP    & \xmark   & 197 & 32   & 62.62 & 35.54 & 45.42 & 24.22 \\ \hline                    
        FLIP    & \xmark   & 49 & 32   & 54.73 & 29.34 & 38.62 & 20.88 \\ \hline
        \multirow{2}{*}{FLIP}   & 
            truncation & \multirow{5}{*}{49} & \multirow{5}{*}{16}   & 44.67 & 25.54 & 34.99 & 19.64 \\
            & random     &  &    & \textbf{58.48} & \textbf{32.36} & 43.61 & \textbf{22.81} \\ \cline{1-2}
        \multirow{2}{*}{CLIPA} 
            & block      &      &    & 56.51 & 30.82 & \textbf{44.06} & 22.66 \\
            & syntax     &      &    & 54.54 & 29.60 & 41.16 & 21.40 \\ \cline{1-2}\cline{5-8}
        CLIPF & 
            frequency &      &     & 57.89 & 31.52 & 42.72 & 22.57 \\ \hline
        \multirow{2}{*}{FLIP}   & 
            truncation   & \multirow{5}{*}{49} & \multirow{5}{*}{8}  & 30.47 & 16.92 & 23.96 & 12.78  \\
            & random     &      &   & 58.58 & \textbf{36.24} & 43.79 & 23.16 \\ \cline{1-2}
        \multirow{2}{*}{CLIPA}
            & block      &      &   & \textbf{60.06} & 35.88 & \textbf{45.98} & \textbf{24.65} \\
            & syntax     &      &   & 50.30 & 29.64 & 38.46 & 20.28 \\ \cline{1-2}\cline{5-8}
        CLIPF & 
            frequency  &      &   & 58.68 & 34.74 & 44.58 & 23.31 \\ \hline
        \multirow{2}{*}{FLIP}   & 
            truncation & \multirow{5}{*}{49} & \multirow{5}{*}{6}   &  32.25 & 16.18 & 22.39 & 11.93 \\
            & random     &      &   & 55.23 & 32.58 & \textbf{42.47} & 21.34 \\ \cline{1-2}
        \multirow{2}{*}{CLIPA}
            & block      &   &   & 54.44 & \textbf{33.12} & 41.68 & \textbf{21.73} \\
            & syntax     &   &   & 46.15 & 26.78 & 34.08 & 17.99 \\ \cline{1-2}\cline{5-8}
        CLIPF & 
        frequency  &   &   & \textbf{56.02} & 32.32 & 41.05 & 21.28 \\ \hline
        \multirow{2}{*}{FLIP}   & 
            truncation & \multirow{5}{*}{49} & \multirow{5}{*}{4}  & 30.97 & 17.44 & 21.09 & 11.07 \\
            & random     &   &  & 41.62 & 24.14 & 30.26 & 15.63 \\ \cline{1-2}
        \multirow{2}{*}{CLIPA}
            & block      &   &  & 40.83 & 23.56 & 30.77 & 14.98 \\
            & syntax     &   &  & 38.66 & 21.68 & 26.08 & 14.14 \\ \cline{1-2}\cline{5-8}
        CLIPF & 
        frequency &   &  & \textbf{45.07} & \textbf{25.60} & \textbf{31.72} & \textbf{15.96} \\ 
        \bottomrule
\end{tabular}
}
\caption{Zero-shot Image-Text retrieval evaluation. We evaluated CLIP, FLIP, CLIPA, and CLIPF image-text retrieval performance on the COCO and Flickr30k datasets. The models are pre-trained on \textbf{CC12M}~\cite{changpinyo2021cc12m} for 30 epochs with image masking (75\%) to speed up training and fine-tune the model an additional epoch without image and text masking.
    }
    \label{tab:zero-shot retrieval}
\end{table}

\begin{table*}[!th]
    \centering
    \resizebox{\textwidth}{!}{%
    \begin{tabular}{ll|c|ccccccccccccccccccccccccc}
    \toprule
    \textbf{Models} & \textbf{Method} & \textbf{\rotatebox{90}{Text Tokens}} & \textbf{\rotatebox{90}{Food-101}} & \textbf{\rotatebox{90}{CIFAR-10}} & \textbf{\rotatebox{90}{CIFAR-100}} & \textbf{\rotatebox{90}{CUB200}} & \textbf{\rotatebox{90}{SUN397}} & \textbf{\rotatebox{90}{Cars}} & \textbf{\rotatebox{90}{Aircraft}} & \textbf{\rotatebox{90}{DTD}} & \textbf{\rotatebox{90}{OxfordPets}} & \textbf{\rotatebox{90}{Caltech-101}} & \textbf{\rotatebox{90}{Kinetics700}} & \textbf{\rotatebox{90}{Flowers102}} & \textbf{\rotatebox{90}{MNIST}} & \textbf{\rotatebox{90}{STL10}} & \textbf{\rotatebox{90}{EuroSAT}} & \textbf{\rotatebox{90}{Resisc45}} & \textbf{\rotatebox{90}{GTSRB}} & \textbf{\rotatebox{90}{KITTI}} & \textbf{\rotatebox{90}{Country211}} & \textbf{\rotatebox{90}{PCAM}} & \textbf{\rotatebox{90}{UCF101}} & \textbf{\rotatebox{90}{CLEVR}} & \textbf{\rotatebox{90}{HatefulMemes}} & \textbf{\rotatebox{90}{SST2}} & \textbf{\rotatebox{90}{ImageNet}} \\ 
    \midrule
    CLIP & \xmark & 32 & 45.46 & 66.42 & 32.45 & 7.90 & 49.27 & 16.17 & 2.28 & 15.48 & 57.51 & 71.91 & 24.84 & 2.12 & 8.77 & 91.29 & 19.08 & 37.38 & 5.09 & 29.81 & 4.45 & 50.05 & 40.52 & 21.84 & 52.66 & 48.76 & 36.61 \\ \hline
    FLIP & \xmark & 32 & 36.26 & 58.09 & 26.89 & 8.73 & 38.56 & 8.16 & 2.98 & 19.95 & 51.85 & 67.17 & 21.57 & 2.94 & 9.35 & 81.89 & 18.02 & 32.30 & \textbf{9.68} & 21.39 & 3.63 & 50.26 & 36.51 & 14.67 & 51.37 & \textbf{50.08} & 33.70 \\ \hline
\multirow{2}{*}{FLIP}   &
    truncation & 16 & 35.53 & 61.24 & 26.54 & 9.42 & 39.98 & 7.28 & 2.57 & 15.80 & 46.85 & 68.28 & 20.47 & 1.94 & 8.32 & 81.89 & 18.48 & 33.75 & 2.33 & 17.98 & 3.79 & 58.56 & 30.85 & 13.31 & 53.00 & \textbf{50.08} & 32.83 \\
    & random & 16 & 38.48 & 53.61 & 29.46 & 7.97 & 41.92 & 8.53 & \textbf{2.79} & 21.54 & \textbf{56.71} & 69.67 & 22.88 & 2.23 & 9.73 & 84.86 & 17.90 & \textbf{36.41} & 7.28 & 32.95 & \textbf{4.77} & 55.00 & 35.29 & 16.38 & 49.93 & \textbf{50.08} & 34.26 \\ \cline{1-2}
\multirow{2}{*}{CLIPA}
    & block & 16 & 38.69 & 57.51 & 29.71 & 8.23 & 44.54 & \textbf{9.85} & 2.45 & 21.33 & 54.01 & 70.61 & 23.00 & 2.79 & 10.09 & \textbf{85.04} & \textbf{31.88} & 29.98 & \textbf{9.07} & 25.07 & 4.59 & \textbf{60.68} & \textbf{38.07} & \textbf{18.72} & 51.24 & 49.64 & 34.78 \\
    & syntax & 16 & 35.94 & 58.78 & 31.86 & 8.91 & 41.92 & 8.39 & 1.53 & 17.07 & 47.05 & 67.59 & 21.34 & \textbf{3.10} & \textbf{10.45} & 76.92 & 26.68 & 32.65 & 8.73 & \textbf{39.37} & 4.23 & 49.96 & 32.91 & 15.55 & \textbf{53.30} & 49.92 & 34.41 \\ \hline
CLIPF & 
    frequency & 16 & \textbf{39.47} & \textbf{64.43} & \textbf{32.55} & \textbf{10.15} & \textbf{45.51} & 9.22 & 2.43 & \textbf{21.70} & 55.62 & \textbf{70.69} & \textbf{23.08} & 2.77 & 9.56 & 83.00 & 21.02 & 33.30 & 7.20 & 27.47 & 4.12 & 50.24 & 37.72 & 13.70 & 52.70 & 49.92 & \textbf{35.96} \\ \hline
\multirow{2}{*}{FLIP}   & 
    truncation & 8 & 36.28 & 61.99 & 32.24 & 8.65 & 44.40 & 5.56 & 2.04 & 18.56 & 45.67 & 65.93 & 22.68 & 2.52 & 10.51 & 87.96 & 34.80 & 29.37 & 3.35 & \textbf{46.06} & 4.59 & 51.55 & 32.38 & 12.41 & 53.46 & 50.08 & 30.30 \\ 
    & random & 8 & 45.10 & \textbf{68.54} & \textbf{34.04} & 9.03 & \textbf{51.54} & 8.64 & 3.23 & \textbf{30.53} & 59.29 & 70.06 & \textbf{27.29} & \textbf{4.21} & 9.82 & 90.46 & 28.72 & 32.54 & 5.99 & 30.61 & \textbf{7.95} & 58.00 & 42.90 & 12.87 & 53.40 & 50.08 & 37.25 \\ \cline{1-2}
\multirow{2}{*}{CLIPA}
    & block & 8 & 43.70 & 66.66 & 32.62 & 8.08 & 51.07 & 9.27 & 2.54 & 28.99 & 57.88 & 70.89 & 26.72 & 2.65 & \textbf{18.10} & \textbf{90.84} & \textbf{31.88} & \textbf{36.94} & \textbf{9.02} & 33.36 & 7.52 & \textbf{61.91} & 40.10 & \textbf{14.66} & 51.80 & \textbf{53.27} & 37.88 \\ 
    & syntax & 8 & 37.92 & 65.25 & 31.72 & 8.28 & 45.74 & 8.73 & 2.96 & 17.07 & 46.54 & 68.73 & 23.52 & 2.59 & 11.30 & 88.30 & 29.90 & 30.81 & 6.47 & 30.82 & 5.02 & 54.62 & 36.58 & 12.51 & 52.74 & 50.08 & 34.97 \\ \hline
CLIPF & 
    frequency & 8 & \textbf{45.75} & 59.00 & 30.76 & \textbf{10.60} & 50.06 & \textbf{10.32} & \textbf{3.20} & 29.89 & \textbf{67.54} & \textbf{73.35} & 26.96 & 4.09 & 10.10 & 90.54 & 20.80 & 36.30 & 7.16 & 40.31 & 7.10 & 57.23 & \textbf{43.09} & 12.27 & \textbf{54.54} & 50.03 & \textbf{39.30} \\ \hline
\multirow{2}{*}{FLIP}   & 
    truncation & 6 & 23.54 & 60.27 & 23.52 & 5.70 & 36.54 & 3.82 & 2.03 & 10.32 & 31.12 & 61.15 & 17.77 & 1.86 & 9.94 & 85.25 & 20.62 & 23.68 & 7.13 & 16.71 & 3.45 & 50.23 & 24.53 & 12.61 & 53.62 & 50.08 & 23.28 \\
    & random & 6 & 38.45 & \textbf{71.57} & 33.89 & 7.73 & 51.81 & 6.67 & 2.72 & 29.47 & 59.92 & 68.65 & 25.89 & 2.42 & \textbf{12.28} & 89.73 & 26.90 & 33.17 & 8.89 & 29.75 & 7.23 & 56.87 & 38.62 & \textbf{12.63} & 51.96 & 50.08 & 34.62 \\ \cline{1-2}
\multirow{2}{*}{CLIPA}
    & block & 6 & 39.87 & 66.60 & \textbf{35.31} & 7.58 & \textbf{53.23} & 7.49 & \textbf{2.89} & 27.87 & 56.74 & \textbf{69.40} & 25.97 & \textbf{2.95} & 10.06 & 89.30 & 27.90 & 32.40 & 7.31 & 33.76 & 6.73 & \textbf{63.10} & 39.20 & 12.65 & \textbf{55.41} & \textbf{52.83} & 35.89 \\
    & syntax & 6 & 34.18 & 63.46 & 29.83 & 7.54 & 43.54 & 7.92 & 1.85 & 17.23 & 44.97 & 68.37 & 22.75 & 1.99 & 11.37 & 89.19 & 17.56 & 26.13 & 3.84 & 33.16 & 4.98 & 55.25 & 35.40 & 12.30 & 53.44 & 50.80 & 32.57 \\ \hline
CLIPF & 
    frequency & 6 & \textbf{45.33} & 67.64 & 32.83 & \textbf{9.98} & 50.86 & \textbf{9.61} & 2.19 & \textbf{31.33} & \textbf{66.03} & 69.24 & \textbf{26.60} & 2.94 & 10.71 & \textbf{90.98} & \textbf{33.64} & \textbf{33.22} & \textbf{8.93} & \textbf{36.70} & \textbf{7.32} & 59.96 & \textbf{40.60} & 12.45 & 54.15 & \textbf{50.08} & \textbf{37.81} \\ \hline
\multirow{2}{*}{FLIP}   & 
    truncation & 4 & 20.64 & 51.67 & 22.10 & 4.19 & 37.16 & 3.11 & 1.18 & 15.16 & 29.36 & 53.99 & 16.17 & 1.04 & 10.10 & 86.34 & 17.16 & 15.81 & 7.48 & 30.15 & 3.49 & \textbf{61.43} & 24.95 & 12.63 & 50.76 & 50.08 & 19.80 \\
    & random & 4 & 28.82 & 61.90 & 28.98 & 5.35 & 46.39 & 4.03 & 1.42 & 25.16 & 43.26 & 62.87 & 22.03 & 2.10 & \textbf{12.24} & 88.54 & 23.82 & 24.46 & \textbf{8.07} & 39.30 & 5.59 & 50.62 & 33.52 & 12.65 & 51.86 & 50.08 & 27.13 \\ \cline{1-2}
\multirow{2}{*}{CLIPA}
    & block & 4 & 28.08 & 60.63 & 27.10 & 5.09 & 45.61 & 3.82 & 2.43 & 23.24 & 36.63 & 64.15 & 21.80 & 2.29 & 8.41 & 87.04 & 24.44 & 23.95 & 7.55 & \textbf{43.05} & 5.51 & 55.80 & 32.62 & 9.49 & \textbf{53.55} & 50.08 & 26.66 \\
    & syntax & 4 & 25.59 & 54.13 & 27.46 & 4.73 & 42.26 & 6.26 & \textbf{2.66} & 15.27 & 32.52 & 58.93 & 19.58 & 2.20 & 10.42 & 90.11 & \textbf{26.02} & 20.62 & 5.19 & 37.50 & 4.32 & 51.12 & 26.82 & \textbf{12.93} & 52.77 & 50.08 & 24.61 \\ \hline
CLIPF & 
    frequency & 4 & \textbf{36.09} & \textbf{69.76} & \textbf{31.14} & \textbf{8.41} & \textbf{47.72} & \textbf{7.01} & 2.24 & \textbf{25.64} & \textbf{54.23} & \textbf{64.55} & \textbf{23.55} & \textbf{2.95} & 10.18 & \textbf{90.25} & 20.78 & \textbf{26.52} & 6.90 & 36.63 & \textbf{6.40} & 55.71 & \textbf{35.71} & 12.65 & 49.63 & 50.08 & \textbf{30.93} \\ 
    \bottomrule
    \end{tabular}}
    \caption{\textbf{Zero-shot accuracy on more classification datasets}. Comparison of the zero-shot accuracy performance of CLIP, FLIP, CLIPA, and CLIPF on various datasets.
        The models are pre-trained on \textbf{CC12M}~\cite{changpinyo2021cc12m} for 30 epochs with image masking (75\%) to speed up training and fine-tune the model an additional epoch without image and text masking.
        }
    \label{tab:dataset_performance}
    \vspace{-0.3cm}
\end{table*}

\subsection{Zero-shot Image-Text Retrieval}
\label{sec:0retrieval}
\cref{tab:zero-shot retrieval} shows image-text retrieval results on Flickr30k~\cite{young2014Flickr30k} and COCO~\cite{lin2014MSCOCO}.
In general, we see that text masking approaches are able to improve over FLIP.
When using 16 text tokens, all strategies except the truncation method outperform FLIP, which is pre-trained on the full text. 
Here, we do not see consistent improvement over CLIP, although some cases come close.
Note again that we are testing here only with 49 image tokens and 98 would fare better (see Tab. 8 of the supplementary material), with additional computational cost.

In contrast to the previous experiments, with image-text retrieval, we observe that CLIPF does not clearly dominate the other text masking strategies.
The CLIPA random, CLIPA block, and CLIPF (frequency) strategies exhibit comparable performance in image and text retrieval across both datasets.
The notable exception is the performance of CLIPF in the case that the text masking ratio is very high, i.e., only 4 input text tokens are used.
Here, the benefit of CLIPF's frequency-based approach is clearly evident.

We point out that for image-text retrieval the performance of syntax masking is particularly low in comparison to the other forms of masking.
Specifically, when using 8 and 6 text tokens, the performance gap between syntax masking and block, random, and frequency masking is approximately 9\% on COCO and 6\% on Flickr30k for image-text retrieval.
This gap is wider than what was observed with zero-shot classification.
We conjecture that the bad performance is due to the image-text retrieval task requiring understanding the overall meaning and logic of the text rather than just recognizing nouns or objects.
For this reason, the fact that syntax masking retains a high number of nouns, as shown in ~\cref{sec:freq}, is particularly harmful.

\subsection{Zero-shot Classification on Additional Datasets}  
\label{sec:add_data}
In ~\cref{tab:dataset_performance}, we present evaluation results 
on additional datasets. 
With six text tokens, CLIPF consistently outperforms the original CLIP (no masking), FLIP (image masking with full-text), and other text masking strategies.
Aligned with ImageNet findings, reducing text tokens from 16 to 8 substantially boosts performance.
Improvements include SUN397~\cite{Xiao10SUN397} (+6.39\%), DTD~\cite{cimpoi14dtd} (+9.31\%), OxfordPets~\cite{parkhi12oxford_pets} (+4.64\%), STL10~\cite{adam11stl10} (+9.31\%), EuroSAT~\cite{helber2019eurosat} (+6.60\%), UCF101~\cite{soomro2012ucf101} (+6.68\%).
The performance of other text masking strategies, including truncation, random, block, and syntax, also improves with this reduction. 
However, their improvements are more limited.

We point out that the Describable Textures Dataset (DTD)\cite{cimpoi14dtd} contains 47 categories of texture patches, whose POS is mostly verb and that the OxfordPets dataset\cite{parkhi12oxford_pets} also contains many class names that are not nouns.
On these data sets, the performance of syntax masking is particularly low.
Specifically, when using six text tokens, syntax masking underperforms CLIPF by 14.10\% on DTD and 21.04\% on OxfordPets.
This low performance reflects that the VLM has overfitted on the large number of nouns retained by syntax masking and mistakes in POS detection (which is challenging for OxfordPets).
Note that, unlike the performance on ImageNet (see Figs. 4 and 5 of the supplementary material), syntax masking on DTD remains worse than random, block, and frequency masking throughout all epochs.
 
\section{Conclusion and Outlook} 

This paper has provided a systematic overview of text masking approaches for VLM training, which can boost performance and reduce computation.
VLMs are important because of their wide use in practical applications, including improving Multimodal Large Language Models (MLLMs)~\cite{li2022blip,liu2023llava}.
We have carried out an analysis of how text masking impacts word frequency in the training data and discovered that text masking strategies perform better if they do not disproportionately impact word frequency.
Our experiments also demonstrate that training epochs have a substantial impact on the performance of different text masking strategies.
We conclude, contrary to the existing state-of-the-art, CLIPA~\cite{li2023clipa}, that syntax masking is not the best performing text masking approach.

Further, this paper has introduced CLIPF, a strategy for text masking based on word frequency.
Our experiment shows that CLIPF yields strong performance and shows the ability to outperform the original CLIP, while using less training resources.
Unlike syntax-based masking methods, CLIPF does not require part-of-speech (POS) information, which consumes more resources to generate than frequency statistics.
CLIPF is also interesting because it seems to balance nouns with words in the ``Other'' POS category, in contrast to syntax masking.

We carry out a quantitative comparison among CLIPF and existing text matching strategies that reveals the strong performance of CLIPF in a large range of zero-shot classification tasks.
CLIPF demonstrates a notable ability to learn image and text representations when compared to a model pre-trained on text masked with other existing text masking strategies.
It also outperforms other approaches under highly aggressive masking (i.e., using only 4 tokens per training text).

This paper has argued that frequency is what you need to think about when choosing a text masking strategy.
Specifically, both truncation and syntax masking have a disproportionate impact on the frequencies of specific words, making clear why they do not perform well.
Further, CLIPF, which exploits word-frequency directly, is a strong performer.
These insights are a valuable guide for researchers and developers, who need to select an appropriate text masking strategy, while taking in to account available computational resources.
Moving forward, CLIPF can be further optimized to more directly shape the  word frequency distribution, 
providing even more control when trading off VLM performance and computation during text masking. 

{
    \small
    \bibliographystyle{ieeenat_fullname}
    \bibliography{main}
}

\clearpage
\appendix
\section{Details of Experimental Setup}

Our implementation follows CLIP~\cite{radford2021clip}, OpenCLIP~\cite{gabriel2021openclip}, FLIP~\cite{li2023flip} and CLIPA~\cite{li2023clipa}.
In this section, we present the details of our experimental setup. 


\textbf{Dataset} 
Data set statistics are summarized in ~\cref{tab:dataset}.
For the majority of our experiments, we pre-train the models on Conceptual Captions 3M (CC3M)~\cite{sharma2018cc3m}, Conceptual 12M (CC12M)~\cite{changpinyo2021cc12m}.
These datasets have been used by OpenCLIP, DeCLIP, SLIP, A-CLIP and BLIP~\cite{gabriel2021openclip,li2022DeCLIP,mu2021slip,yang2023aclip,li2022blip} 
To reproduce CLIPA, we use LAION-400M~\cite{schuhmann2021laion400m}, which is used by CLIPA.
LAION-400M is approximately 133 times larger than CC3M and 32 times larger than CC12M, indicating that it contains a broader range of concepts and offers greater diversity compared to CC3M and CC12M.
 
Note that the difference in caption length with have an impact on masking performance, although we do not directly measure the difference in this paper.

\textbf{Architecture}
For the image encoder, we use ViT-B/16 (86M parameters)~\cite{dosovitskiy2020ViT} architectures with global average pooling and learnable positional embeddings. 
For the text encoder, we use a Transformer-based model~\cite{Guyon2017Transformer} and byte-pair encoding with a 49K token vocabulary.
Additionally, we run one experiment with a larger image encoder (VIT-L/16) with 303M parameters and report the results in the~\cref{sec:more_resutls} of supplementary material.
We chose these encoders because they are the largest ones available to us, given our current resource constraints.
The input image size is 224 for all datasets. 
When using full text for training, the maximum context length is 32.
Zero-padding is applied to input text that is shorter than the maximum token length of the model.

We trained the models using 8 RTX A5000 GPUs with the same settings to ensure consistent conditions across all models.
We experiment with different text token input sizes, namely, 32, 16, 8, 6, and 4 text tokens.
As the input size gets smaller, we can increase the batch size, maximizing computational memory usage.
For CC3M and CC12M, the batch sizes corresponding to the input sizes are: 664, 832, 896, 928, and 944. 
Note that CLIP (and therefore CLIPA and CLIPF) is trained using contrastive learning, which benefits from larger batch sizes.
Note also that although we used different text masking with the same settings and batch sizes, syntax masking was slower than the other text masking strategies when conducting the experiments because POS processing for each word is time-consuming.

We pre-train the model using the InfoNCE loss~\cite{oord2018InfoNCE} with a learnable temperature parameter $\tau$~\cite{chen2020simple,radford2021clip}.
To classify images, we calculate the cosine similarity between the image and text embeddings.

\begin{table}[!t]
    \centering
    \resizebox{\linewidth}{!}{%
    \begin{tabular}{c|c|c|c}
    \toprule
    Dataset & Samples & Total words & Caption length \\   
    \midrule
        CC3M       & $2.72\times10^{6}$   & $2.80\times10^{7}$   & $10.30 \pm 4.72$  \\ 
        CC12M      & $9.30\times10^{6}$   & $2.06\times10^{8}$   & $22.15 \pm 17.20$ \\ 
        LIAON-400M & $2.98\times10^{8}$   & $3.71\times10^{9}$   & $12.51 \pm 15.82$ \\
    \bottomrule
    \end{tabular}
    }
    \caption{Dataset statistics for pre-training datasets. Caption length refers to the number of words in the text.}
    \label{tab:dataset}
\end{table}

\begin{table}[!t]
    \centering
    \resizebox{\linewidth}{!}{%
    \begin{tabular}{l|l|l}
    \toprule
        \textbf{Config} & \textbf{Pre-training} & \textbf{Fine-tuning} \\
        \midrule
        optimizer & AdamW~\cite{loshchilov2017adaW} & AdamW~\cite{loshchilov2017adaW} \\
        learning rate & 1e-3 & 1e-5 \\
        weight decay & 0.2 & 0.2 \\
        optimizer momentum & $\beta_1, \beta_2=0.9, 0.98$~\cite{chen2020generative} & $\beta_1, \beta_2=0.9, 0.98$~\cite{chen2020generative} \\
        learning rate schedule & cosine decay~\cite{loshchilov2016sgdr} & cosine decay~\cite{loshchilov2016sgdr} \\
        warmup steps & 10k & 10\% \\
        epoch & 30 & 1 \\
        $t$ & $10^{-6}$ & --- \\
        $\tau$ & 0.07 & --- \\
        numerical precision & amp & amp \\
        RandomResizedCrop &  (50, 100)  & (50, 100)  \\
        \bottomrule
    \end{tabular}}
    \caption{Details of the pre-training and fine-tuning setups.}
    \label{tab:setup}
\end{table}

\begin{table}[!t]
    \centering
    \resizebox{\linewidth}{!}{
    \begin{tabular}{c|c|c|c|c|r}
    \toprule
    Image masking & Text masking & Image tokens & Text tokens & Total & Percentage \\ 
    \midrule
    0\%                 &  0.00\%            &  196 & 32 & 228 & 100\%      \\ \hline
    \multirow{5}{*}{75\%}              
                        &  0.00\%            &  49 & 32 & 81 & 35.53\% \\
                        &  50.00\%           &  49 & 16 & 65 & 28.51\% \\
                        &  75.00\%           &  49 & 8  & 57  & 25.00\% \\
                        &  81.25\%           &  49 & 6  & 55  & 24.12\% \\
                        &  87.50\%           &  49 & 4  & 53  & 23.25\% \\ 
    \bottomrule
    \end{tabular}
    }
    \caption{The number of image and text tokens that are processed during pre-training (CC3M and CC12M).}
    \label{tab:tokens}
\end{table}

\textbf{Training} Following CLIP, OpenCLIP~\cite{radford2021clip,gabriel2021openclip}, we pre-train our model for 30 epochs on the CC3M and CC12M datasets.
For the LAION-400M dataset, we pre-train the model for 16 epochs, extending CLIPA’s experiments, which were conducted for 6 epochs on the same dataset.
Details of the pre-training configuration are given in~\cref {tab:setup}.

Similar to FLIP, to speed up training, we apply 75\% image masking to the image encoder as the baseline model.
As a result, the speedup is about 4$\times$ compared to training without image masking, while the reduction in the zero-shot performance of ImageNet-1K classification remains within reasonable bounds.

During text masking, we reduce the number of tokens from 32 to 16, 8, 6, and 4. 
To measure the training speed of CLIP, CLIPA, and CLIPF, we compare the number of text tokens processed by each model. 
As shown in ~\cref{tab:tokens}, the number of text tokens processed by each model during pre-training is calculated based on different image and text masking ratios. 
When we pre-trained the model using image masking, the total number of tokens is 81 and the percentage of text tokens is 39.5\%.
When we apply 50\% text masking, the total number of tokens is 65.
Compared to training without text masking, this results in a speed increase of approximately 20\%.
Moreover, when we apply 87.5\% text masking, the total number of tokens is 53, resulting in a speed increase of approximately 34\% compared to training without text masking.

\textbf{Fine-tuning} 
Following FLIP, and CLIPA, in order to bridge the distribution gap between pre-training and evaluation, we fine-tuned the model without images and text masking.
\textcolor{black}{Note that, fine-tuning the models without masking in masking work is focused on reducing the distribution gap and applied sparingly to avoid undoing the tradeoff.}
The details of the fine-tuning configuration are provided in ~\cref{tab:setup}.

\begin{figure}[!t]
    \centering
    \includegraphics[width=1.0\linewidth]{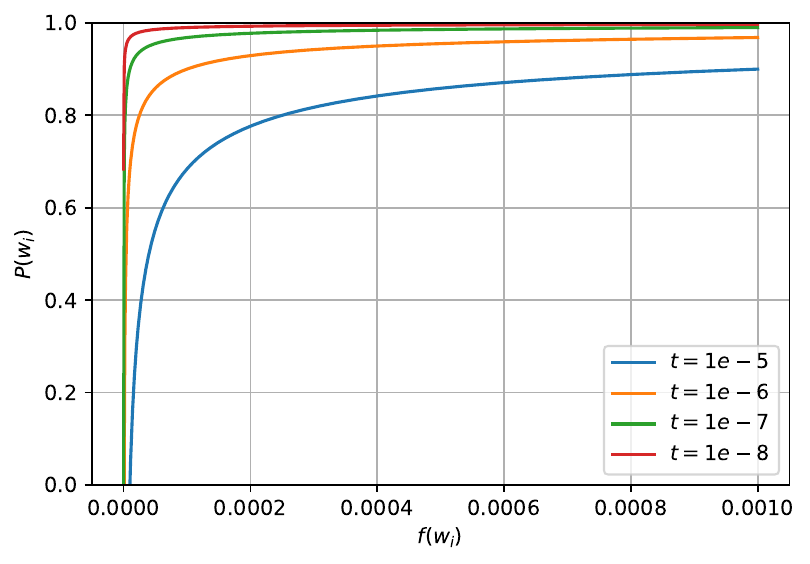}
    \caption{The curve of Equation 2. The x-axis is the word frequency $f(w_i)$, and the y-axis is the $P(w_i)$.
    The value of $t$ of Equation 2 is set to $10^{-5}$, $10^{-6}$, $10^{-7}$,$10^{-8}$.}
    \label{fig:eqution}
\end{figure}

\begin{table}[!t]
    \centering
    \begin{tabular}{l|p{5.5cm}}
    \toprule
    \textbf{Method} & \textbf{Resulting Text} \\ \hline
    original   & Walk of the happy young couple and Siberian dog. The handsome man is hugging the smiling red head girl \\ \midrule
    truncation & walk of the happy young couple \\
    random     & the happy and the is the \\
    block      & couple and siberian dog . the \\
    syntax     & walk dog man smiling head girl \\ \hline
    CLIPF      & siberian handsome man hugging smiling red \\ 
    \bottomrule
    \end{tabular}
    \caption{Example from the CC12M dataset illustrating the effect of various text masking strategies, reducing text length to 6 words. 
    The original caption is in the first row, followed by masked variants. 
    Parameter $t$ in Equation 2 is set to $10^{-6}$.}
    \label{tab:text_masking_cases}
\end{table}

\begin{table}[!t]
    \centering
    \begin{tabular}{l|c}
    \toprule
  Word & Probability \\ \midrule
    walk & 0.926171 \\
    of & 0.992838 \\
    the & 0.995064 \\
    happy & 0.951695 \\
    young & 0.957311 \\
    couple & 0.941174 \\
    and & 0.991695 \\
    siberian & 0.750920 \\
    dog & 0.960531 \\
    . & 0.993678 \\ 
    \bottomrule
    \end{tabular}
\caption{The probability of masking words in the text is calculated using the formula provided in Equation 2.
The example is selected from CC12M~\cite{changpinyo2021cc12m}.
The value of $t$ of Equation 2 is set to $10^{-6}$.}
\label{tab:mask_cases}
\end{table}

\begin{figure*}[!t]
    \centering
    \includegraphics[width=0.9\linewidth]{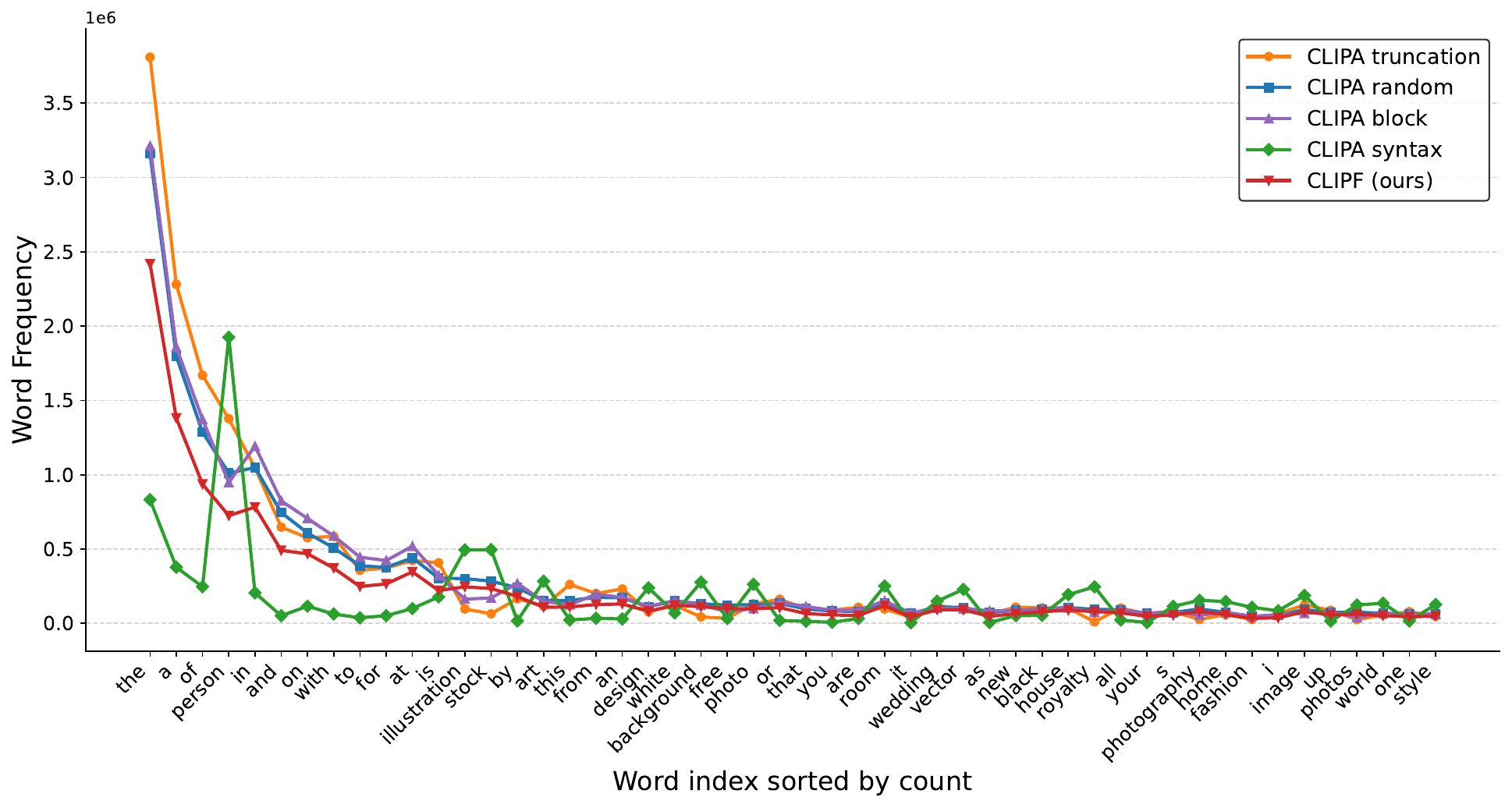}
    \caption{\textbf{The complete x-axis label for Fig. 3. in Sec. 3}
    We set the text length after text masking to 6.
    The x-axis represents the word index, which is sorted by counts of the original data, and the y-axis shows the word frequency.
    The dataset used is CC12M and the value of $t$ of Equation 2 is set to $10^{-6}$.
    We remove special characters from the vocabulary.}
    \label{fig:word_frequency_distribution}
\end{figure*}

\begin{table*}[!t]
    \centering
    \resizebox{\linewidth}{!}{%
    \begin{tabular}{lrrrrrrrrr}
    \toprule
    \textbf{Masking} & \textbf{NN Count} & \textbf{JJ} & \textbf{VB} & \textbf{OTHER} & \textbf{NN (\%)} & \textbf{JJ (\%)} & \textbf{VB (\%)} & \textbf{OTHER (\%)} & \textbf{Total} \\
    \midrule
    Before masking    & 103,469,117 & 10,245,828 & 10,649,943 & 81,351,966  & 50.30\% & 4.98\% & 5.18\% & 39.55\% & 205,716,854 \\ \hline
    Truncation& 28,628,129  & 2,859,462  & 3,272,401  & 20,585,364  & 51.72\% & 5.17\% & 5.91\% & 37.20\% & 55,345,356  \\
    Random    & 28,334,735  & 2,666,847  & 2,790,600  & 21,550,517  & 51.21\% & 4.82\% & 5.04\% & 38.93\% & 55,342,699  \\
    Block     & 27,314,723  & 2,624,594  & 2,897,434  & 22,510,031  & 49.37\% & 4.74\% & 5.24\% & 40.66\% & 55,346,782  \\
    Syntax    & 48,989,317  & 1,483,666  & 1,618,088  & 3,245,892   & 88.53\% & 2.69\% & 2.92\% & 5.87\%  & 55,336,963  \\ \hline
    SW-CLIP       & 34,645,367  & 3,346,676  & 4,073,825  & 6,062,345   & 71.98\% & 6.95\% & 8.47\% & 12.60\% & 48,128,213 \\ \hline
    CLIPF & 33,516,439  & 2,803,409  & 3,473,119  & 15,666,310  & 60.43\% & 5.06\% & 6.20\% & 28.24\% & 55,459,277  \\ \bottomrule
    \end{tabular}
    }
    \caption{Distribution of syntax counts and percentages before and after applying text masking. The dataset is CC12M and we retain 6 words for each text after applying text masking.}
    \label{tab:syntax_percentage_count}
\end{table*}

\textbf{Evaluation setup} 
Following CLIP~\cite{radford2021clip}, FLIP~\cite{li2023flip}, and CLIPA~\cite{li2023clipa}, several classification benchmarks were used.
Among these benchmarks is ImageNet-1K, a widely recognized dataset in computer vision. 
It is frequently used for image classification and VLM tasks and comprises 50K validation samples across 1K different classes.
We filled each class into the templates provided by CLIP~\cite{radford2021clip} to calculate the average of the text embeddings.
We use the same evaluation settings as CLIP~\cite{radford2021clip} to evaluate the other downstream datasets.

\section{Text Masking Analysis}
\subsection{Text Masking Cases}

As illustrated in ~\cref{fig:eqution}, there is a clear relationship between word masking probability and word frequency.
Frequent words have a higher masking probability compared to infrequent ones.
Additionally, a smaller threshold $t$ leads to a smaller difference between the masking probabilities of frequent and infrequent words.
Therefore, it was necessary to choose a relatively larger threshold to ensure that both frequent and infrequent words are effectively masked.

 ~\cref{tab:text_masking_cases} presents the potential text resulting from the text masking technique.
Since truncation is fixed in each epoch, it may result in the loss of important information at the end of the text. Certain words tend to appear in specific positions within the text; for example, “the” and “a” are most likely to be the first words of a sentence. As shown in ~\cref{fig:word_frequency_distribution}, truncation retains more occurrences of “the” and “a” than other text masking strategies.
In contrast, random and block strategies can generate different texts in each epoch for text data augmentation, but they may retain some words with a high frequency.
Syntax masking retains the nouns in the text and remains the same in each epoch. However, this strategy may cause the model to overfit on the frequency of noun words.
In contrast, CLIPF varies the text in each epoch, as words may be retained or removed according to their frequency. This strategy serves two primary purposes: it enhances text diversity and reduces the risk of overfitting to frequent words. 
Another advantage of CLIPF is that it can remove frequent prepositions that are less directly relevant to the objects in the image, such as “a,” “in,” “of,” and others. 
This helps the model focus on the most helpful aspects of the content.
~\cref{tab:mask_cases} shows the masking probabilities for certain words.
High-frequency words such as ``of" ``the" ``and" and ``." are highly likely to be masked from the text.

\subsection{Word Category Distribution Across Different Text Masking Strategies}

As shown in ~\cref{tab:syntax_percentage_count}, we present the word type counts corresponding to Tab. 1 in the main paper.
After applying text masking strategies such as truncation, random, block, syntax, and CLIPF, the word counts remain similar. 
However, SW-CLIP does not maximize the use of the input slots, utilizing only 85\% of the words compared to other text masking strategies.
Additionally, SW-CLIP masks a high percentage of other types of words, which may impact the model’s zero-shot ability.

\begin{figure*}[!t]
    \centering
    \includegraphics[width=0.9\linewidth]{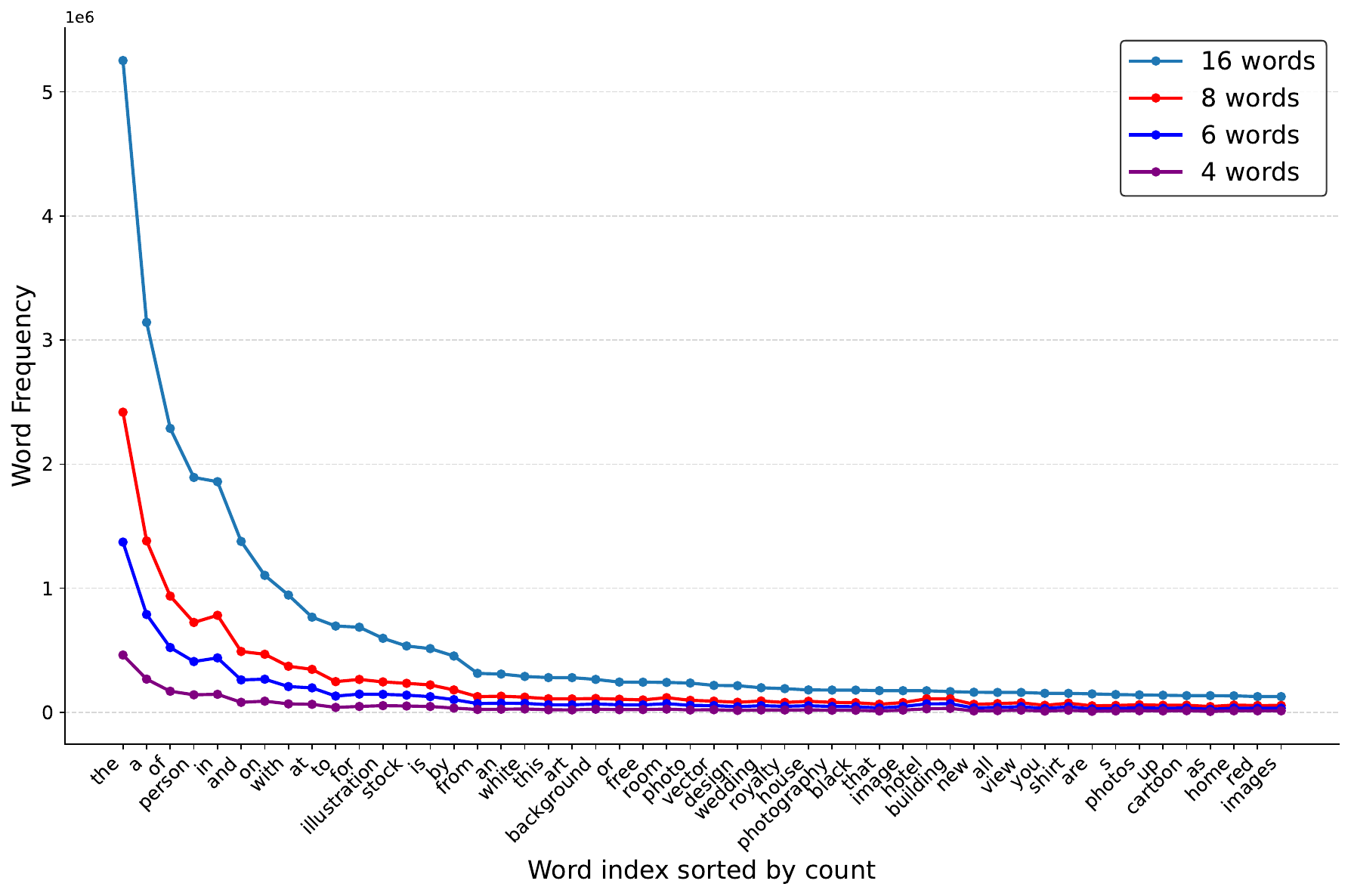}
    \caption{Frequency-based text masking strategies vary according to the number of text words used during pre-training.
    The dataset used is CC12M and the value of $t$ of Equation 2 is set to $10^{-6}$.}
    \label{fig:word_frequency_distribution_frequency}
\end{figure*}

\begin{table*}[!t]
    \centering
    \begin{tabular}{l|l|c|c|cccccc}
    \toprule
    \multirow{2}{*}{Models} & \multirow{2}{*}{Masking} & \multirow{2}{*}{\begin{tabular}[c]{@{}c@{}}Image \\ Tokens\end{tabular}} & \multirow{2}{*}{\begin{tabular}[c]{@{}c@{}}Text \\ Tokens\end{tabular}}
     & \multicolumn{6}{c}{ViT-B/16} \\ 
    & & & & IN-A & IN-O & IN-R & IN-S & IN-V2 & ON \\ 
    \midrule
    CLIP & \xmark & 197 & 32 & 8.97 & \textbf{37.85} & 49.11 & 25.70 & 31.48 & 24.20 \\ \hline
    CLIPF & frequency & 98 & 8 & \textbf{10.37} & 37.75 & \textbf{52.33} & \textbf{28.52} & \textbf{35.62} & 24.00 \\ 
    \bottomrule
    \end{tabular}
    \caption{Zero-shot robustness evaluation. Comparison of the zero-shot accuracy performance of CLIP and CLIPF on various datasets when using more image tokens.
    The models are pre-trained on \textbf{CC12M}~\cite{changpinyo2021cc12m} for 30 epochs with image masking (\textbf{50\%}) to speed up training and fine-tune the model an additional epoch without image and text masking.
    }
    \label{tab:imagenet_more_token}
\end{table*}

\begin{table*}[!t]
    \centering
    \begin{tabular}{l|l|c|c|c|c|c|c}
    \toprule
       \multirow{3}{*}{Models} & \multirow{3}{*}{Masking} & \multirow{3}{*}{\begin{tabular}[c]{@{}c@{}}Image \\ Tokens\end{tabular}} & \multirow{3}{*}{\begin{tabular}[c]{@{}c@{}}Text \\ Tokens\end{tabular}} & \multicolumn{2}{c|}{Text Retrieval}    & \multicolumn{2}{c}{Image Retrieval}    \\
                            & & & & \multicolumn{1}{c|}{Flickr30k} & \multicolumn{1}{c|}{COCO} & \multicolumn{1}{c|}{Flickr30k} & \multicolumn{1}{c}{COCO} \\
                            & & & & R@1 & R@1 & R@1 & R@1 \\
                        \midrule
        CLIP    & \xmark   & 197 & 32   & 62.62 & 35.54 & 45.42 & 24.22 \\ \hline                    
        CLIPF & frequency & 98  & 8  & \textbf{63.11} & \textbf{37.14} & \textbf{46.84} & \textbf{24.89} \\ 
        \bottomrule
\end{tabular}
\caption{Zero-shot Image-Text retrieval evaluation. We evaluated CLIP, CLIPF image-text retrieval performance on the COCO and Flickr30k datasets when using more image tokens.
The models are pre-trained on \textbf{CC12M}~\cite{changpinyo2021cc12m} for 30 epochs with image masking (\textbf{50\%}) to speed up training and fine-tune the model an additional epoch without image and text masking.
}
\label{tab:zero-shot retrieval_more_image_tokens}
\end{table*}

\subsection{Word Distribution Analysis for CLIPF} 

In addition, we analyzed frequency text masking strategies that varied according to the number of words used, as shown in ~\cref{fig:word_frequency_distribution_frequency}.
As the number of words decreased from 16 to 8 and then to 6, more frequent words were masked. 
However, reducing the number to 4 words led to a smaller vocabulary size, resulting in the loss of some important information. 
Consequently, the performance of the model pre-trained with 4 words was substantially lower compared to the model pre-trained with 6 words.
We recommend setting the number of text words in a frequency-based text masking strategy to strike a balance between frequent and infrequent words and to maintain a larger vocabulary.
Based on our experiments, the optimal number of text masking was found to be approximately 40-60\% of the average length of the original text.
This configuration helps achieve a balanced word distribution, which is beneficial for pre-training VLMs.

\begin{table*}[!t]
    \centering
    \resizebox{\linewidth}{!}{%
    \begin{tabular}{l|l|c|c|lll|lll|lll|lll}
    \toprule
       \multirow{3}{*}{Models} & \multirow{3}{*}{Masking} & \multirow{3}{*}{Image Tokens} & \multirow{3}{*}{Text Tokens} & \multicolumn{6}{c|}{Text Retrieval}    & \multicolumn{6}{c}{Image Retrieval}    \\
       & & & & \multicolumn{3}{c|}{Flickr30k} & \multicolumn{3}{c|}{COCO} & \multicolumn{3}{c|}{Flickr30k} & \multicolumn{3}{c}{COCO} \\
       & & & & R@1      & R@5     & R@10     & R@1    & R@5    & R@10   & R@1      & R@5     & R@10     & R@1    & R@5    & R@10   \\
   \midrule
    CLIP    & \xmark   & 197 & 32   & \textbf{62.62} & \textbf{86.00} & \textbf{91.81} & \textbf{35.54} & \textbf{62.38} & \textbf{74.08} & \textbf{45.42} & \textbf{72.56} & \textbf{81.50} & \textbf{24.22} & \textbf{48.42} & \textbf{60.42} \\ \hline     
    FLIP    & \xmark   & \multirow{21}{*}{49} & 32  & 54.73 & 80.37 & 87.97 & 29.34 & 56.08 & 67.00 & 38.62 & 66.09 & 75.40 & 20.88 & 43.22 & 54.78 \\ \cline{1-2} \cline{4-16}
    \multirow{2}{*}{FLIP} & 
truncation &  & \multirow{5}{*}{16}   & 44.67 & 73.08 & 81.85 & 25.54 & 51.90 & 64.64 & 34.99 & 61.05 & 70.85 & 19.64 & 41.91 & 53.51 \\
& random     &  &   & \textbf{58.48} & \textbf{84.62} & \textbf{90.53} & \textbf{32.36} & \textbf{58.76} & 69.98 & 43.61 & 70.67 & 80.04 & \textbf{22.81} & 46.00 & 57.71 \\ \cline{1-2}
\multirow{2}{*}{CLIPA} & 
block    &  &   & 56.51 & 81.26 & 89.05 & 30.82 & 58.32 & \textbf{70.38} & \textbf{44.06} & \textbf{71.20} & \textbf{80.10} & 22.66 & \textbf{46.24} & \textbf{58.06} \\
& syntax     &  &   & 54.54 & 81.07 & 88.36 & 29.60 & 56.52 & 68.54 & 41.16 & 68.32 & 77.51 & 21.40 & 44.82 & 56.56 \\ \cline{1-2} \cline{5-16}
    CLIPF & 
frequency &  &   & 57.89 & \textbf{84.62} & 90.04 & 31.52 & 58.38 & 70.30 & 42.72 & 69.61 & 78.60 & 22.57 & 46.15 & 57.95 \\ \cline{1-2} \cline{4-16}
      \multirow{2}{*}{FLIP} & 
    truncation &  & \multirow{5}{*}{8}  & 30.47 & 59.47 & 70.51 & 16.92 & 38.62 & 51.34 & 23.96 & 47.29 & 57.85 & 12.78 & 31.66 & 42.92 \\
    & random     &  &  & 58.58 & 84.52 & 91.81 & \textbf{36.24} & 62.16 & 72.90 & 43.79 & 70.89 & 80.14 & 23.16 & 46.74 & 58.86 \\  \cline{1-2}
\multirow{2}{*}{CLIPA} & 
    block    &  &  & \textbf{60.06} & 85.01 & \textbf{92.11} & 35.88 & \textbf{62.58} & \textbf{73.74} & \textbf{45.98} & \textbf{73.23} & \textbf{82.49} & \textbf{24.65} & \textbf{48.81} & \textbf{60.70} \\
    & syntax     &  &  & 50.30 & 78.30 & 86.98 & 29.64 & 55.42 & 67.18 & 38.46 & 66.77 & 77.36 & 20.28 & 43.78 & 55.40 \\ \cline{1-2} \cline{5-16}
CLIPF & 
    frequency  &  &  & 58.68 & \textbf{85.10} & 91.51 & 34.74 & 61.38 & 71.88 & 44.57 & 72.58 & 81.92 & 23.30 & 47.50 & 59.24 \\ \cline{1-2} \cline{4-16}
      \multirow{2}{*}{FLIP} &  
truncation &  & \multirow{5}{*}{6}  & 32.25 & 61.05 & 72.98 & 16.18 & 39.52 & 52.48 & 22.39 & 47.46 & 59.39 & 11.93 & 30.03 & 40.86 \\
& random     &  &  & 55.23 & \textbf{82.54} & \textbf{89.25} & 32.58 & 57.54 & 68.68 & \textbf{42.47} & \textbf{70.12} & \textbf{79.43} & 21.34 & 44.79 & 56.26 \\ \cline{1-2}
\multirow{2}{*}{CLIPA} & 
block    &  &  & 54.44 & 79.68 & 88.66 & \textbf{33.12} & \textbf{58.96} & \textbf{70.64} & 41.68 & 69.13 & 79.33 & \textbf{21.73} & \textbf{45.23} & \textbf{57.29} \\
& syntax     &  &  & 46.15 & 76.04 & 84.81 & 26.78 & 52.56 & 64.62 & 34.08 & 61.76 & 73.25 & 17.99 & 40.22 & 52.05 \\ \cline{1-2} \cline{5-16}
    CLIPF & 
    frequency  &  &  & \textbf{56.02} & 82.05 & 88.56 & 32.32 & 58.58 & 70.00 & 41.05 & 69.17 & 78.88 & 21.28 & 44.55 & 56.05 \\ \cline{1-2} \cline{4-16}
\multirow{2}{*}{FLIP} &  
truncation &  & \multirow{5}{*}{4}  & 30.97 & 56.51 & 67.06 & 17.44 & 38.48 & 49.76 & 21.09 & 43.77 & 55.70 & 11.07 & 27.98 & 38.54 \\
& random     &  &  & 41.62 & 69.53 & 80.18 & 24.14 & 48.32 & 60.14 & 30.26 & 56.92 & 68.42 & 15.63 & 35.72 & 47.30 \\ \cline{1-2}
\multirow{2}{*}{CLIPA} & 
block    &  &  & 40.83 & 69.03 & 79.09 & 23.56 & 47.94 & 59.20 & 30.77 & 57.85 & 69.29 & 14.98 & 35.64 & 47.35 \\
& syntax     &  &  & 38.66 & 65.88 & 75.74 & 21.68 & 44.38 & 56.60 & 26.08 & 52.80 & 64.32 & 14.14 & 33.71 & 44.98 \\ \cline{1-2} \cline{5-16}
    CLIPF & 
    frequency &  &  & \textbf{45.07} & \textbf{72.49} & \textbf{81.95} & \textbf{25.60} & \textbf{49.98} & \textbf{61.96} & \textbf{31.72} & \textbf{59.35} & \textbf{71.66} & \textbf{15.96} & \textbf{36.58} & \textbf{48.32} \\ 
    \bottomrule
    \end{tabular}
}
\caption{\textbf{Zero-shot Image-Text Retrieval,} we evaluated CLIP, FLIP, and CLIPF image-text retrieval performance on COCO and Flickr30k datasets.
    The backbone of the image encoder is ViT-B/16, and the model is pre-trained on CC12M for 30 epochs with image masking (75\%) to speed up training and fine-tune the model additional epoch without image and text masking.}
\label{tab:zero_shot_retrieval_more}
\end{table*}

\begin{table*}[!th]
    \centering
    \begin{tabular}{c|c|c|c|c|c}
    \toprule
    \multirow{2}{*}{Models} & \multirow{2}{*}{Datasets} & \multirow{2}{*}{Image Tokens} & \multirow{2}{*}{Text Tokens} & \multicolumn{2}{c}{ViT-B/16} \\ 
     & & & & pre-train & fine-tune \\ \hline
    SW-CLIP & \multirow{2}{*}{CC3M} & \multirow{2}{*}{49} & \multirow{2}{*}{6} & \textbf{14.9} & 16.8 \\
    CLIPF & & & & 14.4 & \textbf{18.2} \\ \hline
    SW-CLIP & \multirow{2}{*}{CC12M} & \multirow{2}{*}{49} & \multirow{2}{*}{8} & 35.9 & 38.5 \\
    CLIPF & & & & \textbf{36.6} & \textbf{39.3} \\ \bottomrule
    \end{tabular}
\caption{Comparison of the zero-shot accuracy performance of SW-CLIP and CLIPF on ImageNet-1k.
    The models are pre-trained on \textbf{CC12M}~\cite{changpinyo2021cc12m} for 30 epochs with image masking (75\%) to speed up training and fine-tune the model an additional epoch without image and text masking.}
\label{tab:swclip}
\end{table*}

\section{More Results}
\label{sec:more_resutls}
More detailed results are presented in this section, including the results and learn curve of image-text retrieval.

\subsection{More image tokens}

As shown in~\cref{tab:imagenet_more_token} and~\cref{tab:zero-shot retrieval_more_image_tokens}, CLIPF achieves better zero-shot robustness in 4 out of 6 datasets and image–text retrieval performance in both Flick30k and COCO datasets than the original CLIP, even though it only uses 50\% of the image tokens and 25\% of the text tokens.

\subsection{The details of Image-text Retrieval}
In~\cref{tab:zero_shot_retrieval_more}, we show more details of Zero-shot Image-Text Retrieval on COCO and Flickr30k datasets.

\begin{figure*}[!t]
    \centering
    \includegraphics[width=0.9\linewidth]{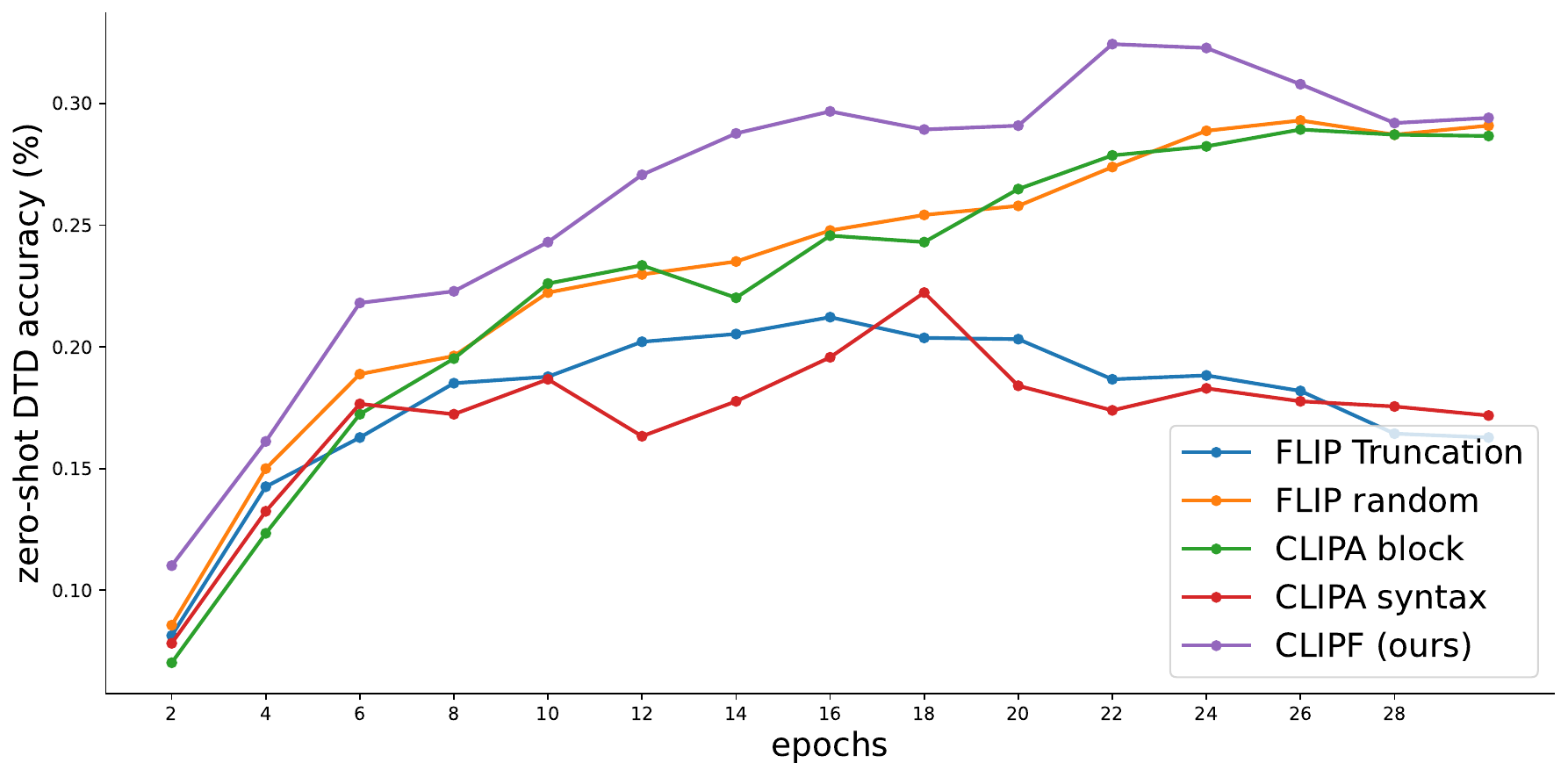}
    \caption{Zero-shot classification accuracy on DTD dataset over training epochs for CLIPF and CLIPA strategies.
    The backbone of the image encoder is ViT-B/16, and the model is pre-trained on CC12M for 30 epochs with image and text masking (75\%) to speed up training and fine-tune the model additional epoch without image and text masking.}
    \label{fig:dtd_curved}
\end{figure*}

\begin{figure*}[!t]
    \centering
    \includegraphics[width=0.9\linewidth]{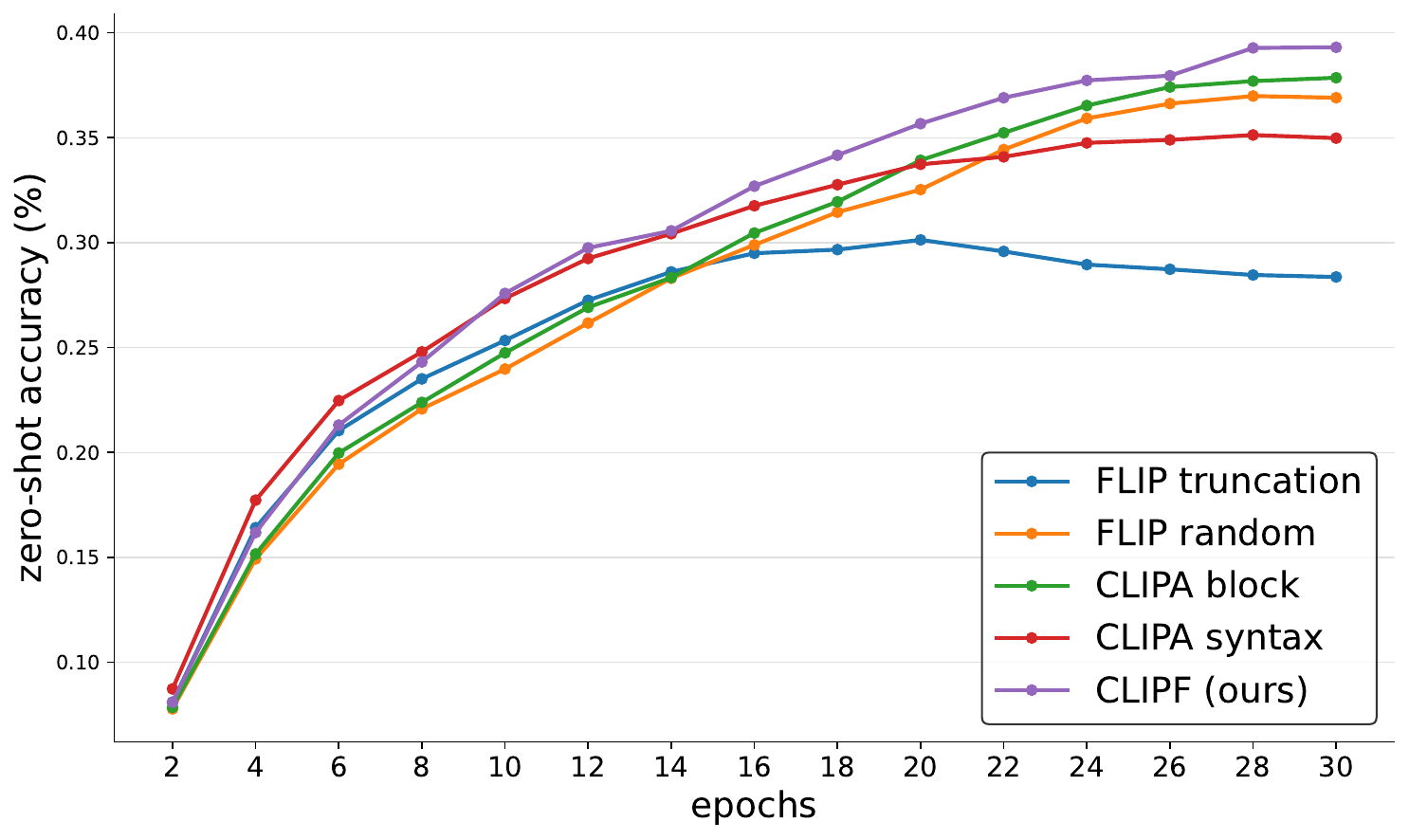}
    \caption{Zero-shot classification accuracy on ImageNet-1K dataset over training epochs for CLIPF and CLIPA strategies \textbf{after fine-tuning}.
     The backbone of the image encoder is ViT-B/16, and the model is pre-trained on CC12M for 30 epochs with image and text masking (75\%) to speed up training and fine-tune the model additional epoch without image and text masking.}
    \label{fig:imagenet}
\end{figure*}

\begin{figure*}[!th]
    \centering
    \includegraphics[width=0.9\linewidth]{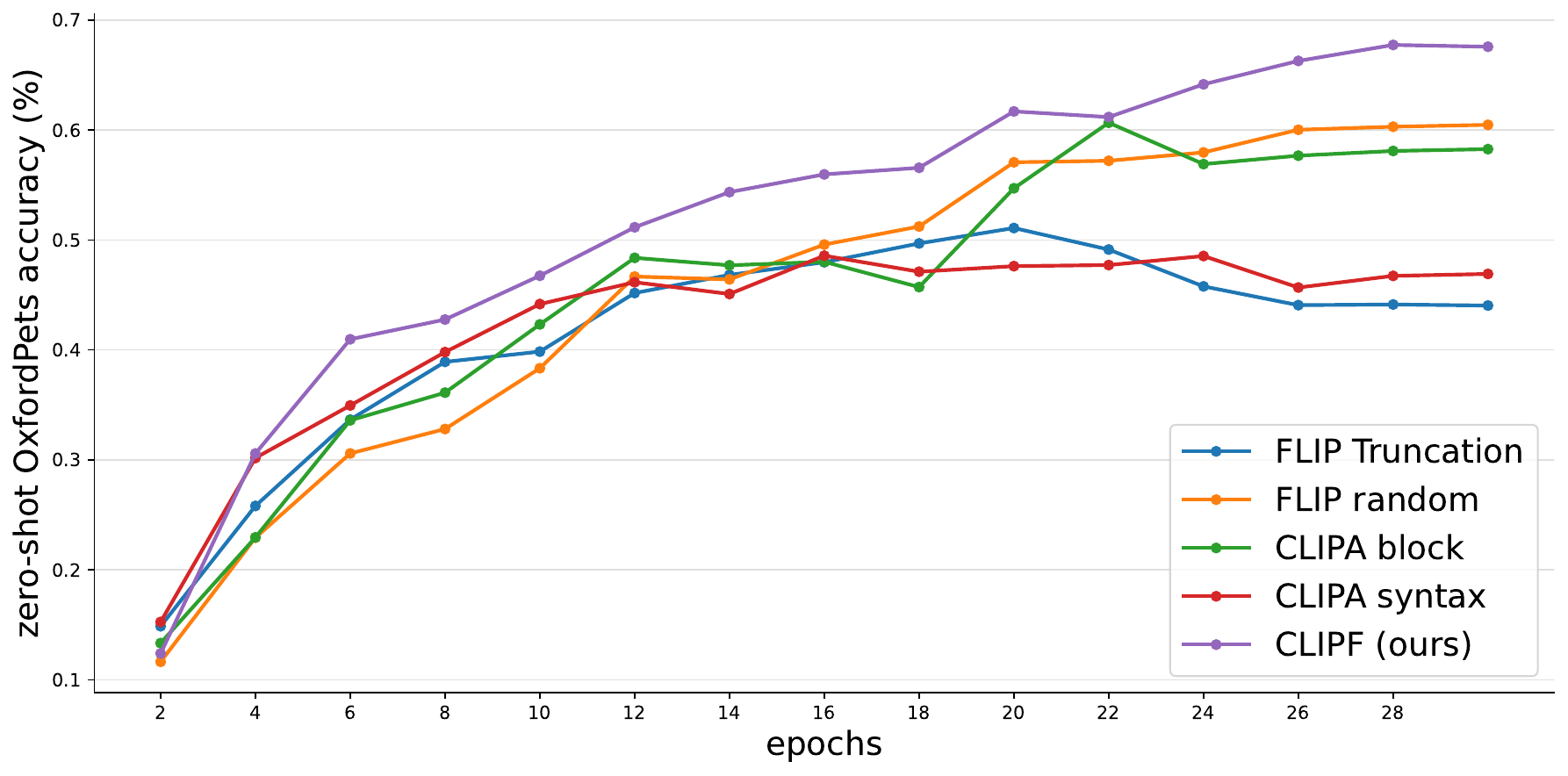}
    \caption{Zero-shot classification accuracy on OxfordPets dataset over training epochs for CLIPF and CLIPA strategies.
    The backbone of the image encoder is ViT-B/16, and the model is pre-trained on CC12M for 30 epochs with image and text masking (75\%) to speed up training and fine-tune the model additional epoch without image and text masking.}
    \label{fig:pets_curved} 
\end{figure*}

\begin{table*}[!th]
\centering
\begin{tabular}{l|l|c|c|cc}
\toprule
    \multirow{2}{*}{Models} & \multirow{2}{*}{Masking} & \multirow{2}{*}{Image  Tokens} & \multirow{2}{*}{Text  Tokens} & \multicolumn{2}{c}{\textbf{ViT-L/16}}  \\
      &    &   &  & pre-train  & fine-tune  \\ \midrule
    \multirow{2}{*}{FLIP}  & truncation      & \multirow{5}{*}{49}     & \multirow{5}{*}{8}     & 28.4 & 31.7 \\
          & random &   &  & 36.0 & 38.3 \\ \cline{1-2}
    \multirow{2}{*}{CLIPA}
          & block &   &  & 36.8 & 39.5 \\
          & syntax &   &  & 32.4 & 37.5\\ \cline{1-2} \cline{5-6}
   CLIPF  & frequency &   &  & \textbf{38.0} & \textbf{40.2} \\ \bottomrule
\end{tabular}
\caption{
Comparison of the ImageNet-1k zero-shot accuracy performance of \textbf{ViT-L/16} with different text masking strategies.
The models are pre-trained on \textbf{CC12M}~\cite{changpinyo2021cc12m} for 30 epochs with image masking (75\%) to speed up training and fine-tune the model an additional epoch without image and text masking.}
\label{tab:vit_l_16}
\end{table*}

\subsection{Learn Curved for DTD and OxfordPets Datasets}

The learning curves for the DTD~\cite{cimpoi14dtd} and OxfordPets~\cite{parkhi12oxford_pets} datasets are presented because not all class names in these datasets are nouns.
As shown in~\cref{fig:dtd_curved}, the performance of syntax masking on the DTD dataset is consistently lower than that of random, block, and frequency masking across all epochs, rather than initially outperforming them and declining at later stages in the ImageNet dataset, as shown in ~\cref{fig:imagenet}.
Moreover, the performance of syntax masking is very poor and almost close to truncation masking.
OxfordPets is a dataset containing incomplete noun class names. 
As shown in~\cref{fig:pets_curved}, in the early training epochs, syntax masking performs similarly to frequency masking and still outperforms truncation, block, and random masking.
However, in the later stages of training, syntax masking performs much worse than the other text masking strategies.

\subsection{Comparison with SW-CLIP}
In this section, we provide a comparison of CLIPF with SW-CLIP~\cite{liang2023swclip}, which also uses frequency-based sampling but imposes a threshold on the frequency score.
The effect of this threshold is that input tokens will go unused if not enough words in a training text surpass the threshold.
In contrast, CLIPF calculates the word frequency by the threshold to prioritize words and maximize the input slots.
For instance, using CC3M as an example, the average length of text before masking is 10.31 ± 4.7, and after SW-CLIPF masking, it is 4.26 ± 2.6~\cite{liang2023swclip}.
It is clear that the use of the input slots is not maximized when the input length of text tokens is set to 6 or longer.
For a fair comparison with CLIPF, we pre-train SW-CLIP using the same setup: 75\% image masking and 81.25\% text masking.
Subsequently, we fine-tuned both models with an additional epoch without any image or text masking. 
The results in~\cref{tab:swclip} show that after fine-tuning CLIPF outperforms SW-CLIP. 

\subsection{Large Image Encoder Architecture}

We carried out an extra experiment which is apply different text masking strategies on larger architectures which is ViT-L/16.
The behavior of different architectures is very consistent when applying different 
text masking strategies, as shown in~\cref{tab:vit_l_16}.

\subsection{Applying text masking on SigLIP}

\textcolor{black}{
We also applied the text masking strategies to SigLIP; the results are presented in ~\cref{tab:sigclip}. CLIPF still achieves superior performance compared to other text masking strategies and without text masking.
This further supports our conclusion that frequency-based masking is the more effective strategy.
}

\begin{table*}[!th]
\centering
\begin{tabular}{l|l|c|c|cc}
\toprule
    \multirow{2}{*}{Models} & \multirow{2}{*}{Masking} & \multirow{2}{*}{Image  Tokens} & \multirow{2}{*}{Text  Tokens} & \multicolumn{2}{c}{ViT-B/16}  \\
      &    &   &  & pre-train  & fine-tune  \\ \midrule
    SigLIP  & \xmark    & 197        & 32        & 39.3& \xmark \\ \hline
    SigLIP  & \xmark    & 49& 32        & 28.4& 29.6\\ \hline
    \multirow{4}{*}{SigLIP}  & truncation      & \multirow{5}{*}{49}     & \multirow{5}{*}{8}     & 20.4& 27.1\\
          & random &   &  & 29.7& 31.7\\
          & block &   &  & 30.5& 32.7\\
          & syntax &   &  & 25.9& 31.6\\ \cline{1-2} \cline{5-6}
   SigLIP  & frequency &   &  & 32.4& 34.8 \\ \bottomrule
\end{tabular}
\caption{
\textbf{Comparison of the ImageNet-1k zero-shot accuracy performance of SigLIP with different text masking strategies.}
The models are pre-trained on \textbf{CC12M}~\cite{changpinyo2021cc12m} for 30 epochs with image masking (75\%) to speed up training and fine-tune the model an additional epoch without image and text masking.}
\label{tab:sigclip}
\end{table*}

\section{Ablation}

\begin{table*}[!t]
    \centering
    \begin{tabular}{c|c|c|c|cc}
    \toprule
    \multirow{2}{*}{model} & \multirow{2}{*}{Image tokens} & \multirow{2}{*}{Text tokens} & \multirow{2}{*}{Threshold} & \multicolumn{2}{c}{ViT-B/16} \\
                           &                               &                              &                            & pre-train     & fine-tune    \\ \midrule
    \multirow{3}{*}{CLIPF} &  \multirow{3}{*}{49} & \multirow{3}{*}{8}  
                            & 1e-5 & 35.6 & 38.4         \\ 
                      &   & & 1e-6 & \textbf{36.6} & \textbf{39.3}   \\ 
                      &   & & 1e-7 & 36.1 & 38.6   \\ \bottomrule
    \end{tabular}
\caption{We pre-train CLIPF on the CC12M dataset across different thresholds in Equation 2.
The models are pre-trained on \textbf{CC12M}~\cite{changpinyo2021cc12m} for 30 epochs with image masking (75\%) to speed up training and fine-tune the model an additional epoch without image and text masking.
}
\label{tab:t}
\end{table*}

\begin{table*}[!t]
    \centering
    \begin{tabular}{c|c|c|c|c|cc}
    \toprule
    \multirow{2}{*}{Models} & \multirow{2}{*}{Masking} & \multirow{2}{*}{Image Tokens} & \multirow{2}{*}{Text Tokens} &  \multicolumn{2}{c}{ViT-B/16} \\ 
       & &  & & pre-train & fine-tune \\ \midrule
    \multirow{2}{*}{CLIPF}   
    & frequency-token   & 49 &  8  & 36.0 & 38.0 \\ 
    & frequency-word    & 49 &  8  & 36.6 & \textbf{39.3} \\ \bottomrule
    \end{tabular}
    \caption{\textbf{Comparison of CLIPF pre-trained with token masking and word masking on ImageNet-1K for zero-shot classification.}
    The models are pre-trained on \textbf{CC12M}~\cite{changpinyo2021cc12m} for 30 epochs with image masking (75\%) to speed up training and fine-tune the model an additional epoch without image and text masking.
    }
    \label{tab:text_word_masking}
\end{table*}

\begin{table}[!th]
  \centering
  \begin{tabular}{l|l|r}
    \toprule
    Model  & Masking & Time (minutes) \\ 
    \midrule
    \multirow{2}{*}{FLIP}   
        & trancation   & 39.81          \\
        & random       & 48.43          \\ \hline
    \multirow{2}{*}{CLIPA}  
        & block        & 40.81          \\
        & syntax       & 217.76         \\ \hline
    CLIPF  & frequency    & 84.64          \\
    \bottomrule
  \end{tabular}
  \caption{Processing times for the CC12M dataset by using different masking strategies.}
  \label{tab:caption-times}
\end{table}

\subsection{Threshold Analysis}
To investigate the impact of the threshold $ t $ in Equation 2 on model performance, we pre-trained models with varying thresholds. 
As shown in~\cref{tab:t}, thresholds of $ 1e-5 $, $ 1e-6 $, and $ 1e-7 $ yield comparable results, all outperforming the other text masking strategies. 
This indicates that CLIPF is relatively insensitive to small variations in thresholds, which are intended to maintain the word masking probability within the range of 0 to 1.
However, using too small a threshold reduces the differences in text masking probabilities; therefore, we do not recommend using an excessively low threshold.

\subsection{Word and Token Analysis}
Since Open\_CLIP encodes text using byte-pair encoding (BPE)~\cite{gabriel2021openclip}, some words are represented by multiple tokens. 
Therefore, in this experiment, we calculate the masking probability based on token frequency rather than whole words.
~\cref{tab:text_word_masking} compares the performance of both approaches. 
Although token frequency masking may disrupt the original word structure and thus achieve lower performance compared to word frequency masking, 
it still slightly outperforms other text masking strategies.

\section{Computational overhead}

We carried out a simple measurement of how long it takes to tokenize CC12M using one thread. 
As shown in ~\cref{tab:caption-times}, syntax masking required 2.5 times the number of minutes of CLIPF (218 vs. 85) and about 5.4 times that of truncation, random, and block masking.

\end{document}